# Self-Supervised Evolutionary Learning of Neurodynamic Progression and Identity Manifolds from EEG During Safety-Critical Decision Making

**Xiaoshan Zhou, S.M. ASCE[1], Carol C. Menassa, F. ASCE[2*], and Vineet R. Kamat, F. ASCE[3]**

[1]Ph.D. Candidate, Dept. of Civil and Environmental Engineering, University of Michigan, Ann Arbor, MI, 48109-2125. Email: xszhou@umich.edu

[2]Professor, Dept. of Civil and Environmental Engineering, University of Michigan, Ann Arbor, MI, 48109-2125. Email: menassa@umich.edu

[3]Professor, Dept. of Civil and Environmental Engineering, University of Michigan, Ann Arbor, MI, 48109-2125. Email: vkamat@umich.edu

\* Corresponding author

**Abstract**

Human–vehicle interaction in safety-critical traffic environments increasingly incorporates neural sensing to infer user intent and cognitive state, yet most existing approaches either treat electroencephalography (EEG) as a static biometric credential or train task-specific decoders that ignore long-term neurodynamic trajectories. These paradigms lack mechanisms to jointly achieve cybersecure user identity and continual modeling of evolving cognitive states during interaction. This work proposes a self-supervised evolutionary learning (SSEL) framework that discovers individualized neurodynamic progressions and intrinsic identity manifolds directly from continuous EEG, without external labels or predefined cognitive stage models. SSEL jointly optimizes within-stage temporal predictability, boundary contrast, cross-trial alignment, and sparse stage-specific feature weights, while a population-based evolutionary search enables direct optimization in the discrete, non-differentiable space of candidate segmentations. We validate the framework on EEG recorded from participants performing a simulated road-crossing decision task, a canonical safety-critical scenario in which perceptual assessment, risk evaluation, and decision commitment unfold over time. The learned segmentations reveal stable, person-specific stage structures and neurodynamic signatures that support authentication and anomaly detection, while maintaining consistency across heterogeneous traffic scenarios. Compared to inference-based temporal segmentation baselines, including Bayesian online change-point detection, kernel-based inference, and Pruned Exact Linear Time, SSEL achieves orders-of-magnitude higher boundary contrast, substantial gains in cross-trial generalization of intention boundaries, and more interpretable, sparse stage-wise feature attributions. Beyond performance, the framework advances a progression-aware perspective on cognitive neurodynamics, where security, resilience, and personalization emerge from the intrinsic temporal structure of brain activity, with implications for next-generation smart urban and transportation infrastructures.

# 1 Introduction

Cognition unfolds in time. Perception, evaluation, deliberation, and action are not instantaneous events, but structured neurodynamic processes that evolve through latent trajectories in high-dimensional neural state space (Wang, 2012). Across a wide range of tasks, such as perceptual discrimination, working memory, motor planning, and decision making, the brain transitions through relatively stable yet transient regimes often conceptualized as stages, states, or metastable attractors (Tognoli & Kelso, 2014). Understanding how such stages emerge, how they are organized temporally, and how they vary across individuals remains a central challenge in cognitive neuroscience (Luo et al., 2025; Parr et al., 2024).

Electroencephalography (EEG), with its millisecond-scale temporal resolution, provides a powerful window into these evolving neurodynamic processes (Jorge et al., 2014). Yet most EEG analysis paradigms remain fundamentally static in orientation. Classical approaches detect condition-dependent differences between predefined intervals (Mahini et al., 2020), while modern machine learning methods decode externally labeled states such as motor imagery, emotional category, or task condition (Weng et al., 2025). Even when sophisticated deep learning architectures are employed, the focus is typically on classifying predefined labels or predicting behavioral outcomes (Liu et al., 2025; Saibene et al., 2024), rather than discovering the intrinsic temporal structure of neural activity itself.

This creates a methodological gap. If cognition is inherently progressive and stage-like, then computational models should be capable of uncovering latent stage boundaries and individualized neurodynamic trajectories directly from continuous neural time series without relying on externally imposed annotations. However, such latent structures are not directly observable. They are embedded within noisy, high-dimensional EEG signals and must be inferred from the internal regularities of temporal dynamics rather than prescribed through behavioral labels.

Self-supervised learning (SSL) offers a principled alternative to traditional supervised paradigms by deriving learning signals from the structure of the data itself. In domains such as vision and language, SSL has demonstrated remarkable success in learning transferable representations through contrastive, predictive, or reconstruction-based objectives (Zhang et al., 2025a). More recently, SSL has been applied to EEG to reduce dependence on labeled data. Nevertheless, most EEG-SSL approaches remain downstream-oriented (Deldari et al., 2022): representations are evaluated by their ability to improve classification of externally defined tasks, rather than by their capacity to reveal the intrinsic neurodynamic organization of cognition.

In particular, existing frameworks rarely treat temporal segmentation as the primary object of inference. Instead, segmentation, if performed at all, is window-based, threshold-driven, or derived from statistical change-point detection methods that test for distributional shifts (Aminikhanghahi & Cook, 2017; Truong et al., 2020). Such inference-based approaches can identify local changes but do not construct an internal model of progression. Similarly, supervised and pretrain–finetune pipelines depend on externally defined stage boundaries, limiting their ability to uncover individualized cognitive trajectories.

A progression-aware perspective requires rethinking the problem formulation. Rather than asking which predefined state a signal belongs to, we ask: how does neural activity organize itself over time? Can we discover temporally coherent stages characterized by internally predictable dynamics, sharp transitions between processing regimes, and reproducible structure across repeated trials? Can such discovered structures reveal stable, person-specific neurodynamic manifolds that reflect how individuals habitually process information and commit to decisions?

To address these questions, we propose a self-supervised evolutionary learning (SSEL) framework for discovering individualized neurodynamic progression directly from continuous EEG. The framework jointly optimizes multiple criteria grounded in cognitive plausibility: within-stage temporal predictability (reflecting metastable state stability), boundary contrast (reflecting rapid transitions between processing regimes), cross-trial alignment (reflecting invariant stage sequences across repetitions), and sparsity (supporting interpretable stage-wise neural signatures). Crucially, because temporal boundaries are discrete and non-differentiable, we adopt a population-based evolutionary search paradigm that directly explores the combinatorial space of candidate segmentations without requiring differentiable relaxations.

We evaluate the proposed framework on EEG recorded during a simulated road-crossing decision task. This paradigm provides a canonical example of safety-critical sequential decision making in which perceptual appraisal, risk evaluation, deliberation, and motor commitment unfold over time. By applying SSEL to this setting, we demonstrate that meaningful stage structures and interpretable channel–band signatures emerge endogenously from neural dynamics, without externally defined cognitive models. Importantly, the proposed methodology is task-agnostic and assumes only that cognition exhibits temporally organized regimes, a property shared by many sequential processes including evidence accumulation, working memory updating, metacognitive evaluation, and multi-step reasoning. The road-crossing scenario therefore serves as an instantiation of a broader problem: uncovering the latent progression of cognitive neurodynamics from continuous brain activity.

The remainder of this paper is organized as follows. Section 2 reviews related work on EEG security, SSL, and temporal segmentation from a cognitive neurodynamic perspective. Section 3 presents the proposed SSEL framework and experimental design. Section 4 reports quantitative results, neurophysiological plausibility analyses, and evolutionary search dynamics. Section 5 discusses the cognitive mechanisms, theoretical implications, and broader generality of progression-aware neurodynamic modeling.

## 2 Literature Review

### 2.1 EEG Security and Privacy: From Biometric Protection to Neurodynamic Security

Research at the intersection of EEG and cybersecurity can be broadly categorized into three lines of work, each addressing a different aspect of security but sharing a common assumption that EEG is either a static credential or a data object to be protected.

The first line treats EEG as a biometric modality for authentication and identification (Baraku et al., 2024; Stergiadis et al., 2022). EEG is considered as a hard-to-forge biometric token, analogous to fingerprints or facial images. Typical systems acquire EEG under either resting or stimulus-evoked paradigms, such as eyes-open/eyes-closed resting state, motor imagery, P300 oddball tasks, or visually evoked potentials. From these recordings, spectral, time–frequency, or spatial–temporal features are extracted and used with traditional classifiers (e.g., LDA, SVM) or deep architectures (e.g., CNNs, RNNs) to perform user identification or verification (Alahaideb et al., 2025; Chan et al., 2018; Khalil et al., 2024). Although promising accuracy has been reported in small-scale, within-session studies, substantial degradation across sessions, days, and contexts remains a persistent challenge (Rehman et al., 2025). Practical limitations related to wearability, setup time, user comfort, and susceptibility to artifacts are also reported (Jalaly Bidgoly et al., 2020). Recent research often positions EEG-based authentication as a secondary or multi-factor modality in high-security contexts such as virtual reality systems, military applications, or critical infrastructure environments (Tucci et al., 2025; Vadher et al., 2024). From a security perspective, however, this line of work conceptualizes EEG primarily as a biometric credential, rather than addressing the security of EEG signals themselves.

A closely related line of research focuses on securing EEG as sensitive data. Because EEG can reveal highly private attributes, such as neurological disorders, emotional states and cognitive capacity, researchers increasingly recognize that using EEG for authentication or decoding introduces new privacy risks. This has motivated work on template protection, anonymization, and synthetic data EEG generation. One major direction is the design of revocable (cancellable) biometric templates, in which non-invertible transformations map EEG feature into protected spaces that can be revoked and reissued even if leaked (Wang et al., 2022b). Another direction is using GANs to generate realistic synthetic EEG for data sharing or training without disclosing real subject data (Debie et al., 2020; Zhou & Liao, 2022). Some studies explicitly introduce differential privacy constraints into GANs to control identifiability (Habashi et al., 2023). There are also broader privacy-preserving brain-computer interface (BCI) frameworks that further examine cross-subject variability, privacy–utility trade-offs, and privacy evaluation metrics, emphasizing the need for unifying privacy protection at the architectural and algorithmic levels (Xia et al., 2022). These researches recognize EEG as a new source of privacy leakage, but they treat EEG largely as static sensitive data to be protected, rather than as a temporally structured signal whose dynamics could themselves contribute to security. As a result, current methods struggle to internalize rich temporal structure or leverage neurodynamic patterns as endogenous security resources.

A third line of work investigates adversarial attacks and defenses for EEG-based BCIs. Recent studies demonstrate that EEG-based BCIs are highly vulnerable to adversarial perturbations, where imperceptible signal modifications can induce severe misclassification in motor imagery or BCI spelling tasks, exposing EEG-BCI systems themselves as a new attack surface (Zhang et al., 2020). Other works explores data poisoning, backdoor, and filtering attacks. Some work proposes narrow-pulse data poisoning and adversarial filtering attacks, where manipulation of training data or online data can significantly degrade decoding performance or hijack outputs under specific trigger conditions (Meng et al., 2024; Meng et al., 2023). In response. emerging security evaluation and defensive benchmarks, such as attention-based defenses, alignment-based defenses, and adversarial training combined with active learning, are. beginning to form an attack–defense ecosystem for EEG-BCI security (Aissa et al., 2024; Chen et al., 2025; Chen et al., 2024; Wu et al., 2023).This body of research conveys a critical insight: EEG is not inherently secure, and when neural signals are directly coupled to control channels, robustness against adversarial attacks must be taken seriously.

Across these three lines of work, research primarily focuses on using EEG as a more secure password, preventing EEG leakage, or improving robustness of EEG classifiers under attack. None explicitly model the neurodynamic progression of cognition or leverage individualized temporal trajectories as intrinsic identity or security markers. This gap motivates a shift from protecting EEG as data toward learning security-relevant structure from neurodynamics themselves.

## 2.2 Self-Supervised Learning for EEG and Temporal Segmentation

SSL has recently attracted increasing attention in EEG research as a means to reduce reliance on costly labeled data (Rafiei et al., 2022; Weng et al., 2025). Most existing EEG–SSL approaches formulate pretext tasks to learn a "high-quality" latent representation from unlabeled EEG, which is subsequently fine-tuned or evaluated on downstream supervised tasks such as sleep staging, emotion recognition, or motor imagery classification (Weng et al., 2025). The dominant paradigm follows a two-stage pipeline: self-supervised pretraining using objectives such as contrastive learning, reconstruction, or prediction, followed by downstream classification or regression with limited labels. Under this paradigm, although SSL is used to learn latent features, segmentation is treated only as a possible downstream task (and is rarely explored). This orientation is also reflected in evaluation practices. Existing EEG–SSL methods are predominantly assessed using downstream task performance like decoding accuracy (Ou et al., 2022). Consequently, SSL is used primarily to improve label efficiency and representation transferability, rather than to infer intrinsic temporal structure. Nevertheless, our work frames segmentation itself as the core inference problem: the objective is to discover temporally coherent cognitive stages or individual-specific neurodynamic trajectories directly from continuous EEG, with the intrinsic temporal structure serving as the supervision signal.

Most EEG–SSL methods derive their self-supervised signals from artificially constructed transformations or contrast structures. Representative categories include contrastive learning (Hu et al., 2024), e.g., SimCLR (Mohsenvand et al., 2020), BENDR (Kostas et al., 2021), ContraWR (Yang et al., 2023), CL-SSTER (Shen et al., 2024), reconstruction or masking-based objectives, e.g., MAEEG (Chien et al., 2022), AFTA-Transformer (Huang et al., 2025), EEGPT (Wang et al., 2024), EEG-TPSR (Dai et al., 2025), and preset transformation recognition tasks such as temporal reordering, channel shuffling, or frequency perturbation, as seen in GMSS (Li et al., 2022). While effective for learning general-purpose embeddings, these approaches do not explicitly target real cognitive boundaries or event segmentation in EEG, as the supervision signal originates from synthetic perturbations rather than endogenous neural dynamics (Weng et al., 2025).

Although SSL-based segmentation remains rare in EEG research, closely related problems have been extensively explored in other temporal modalities, particularly speech and video. In speech and audio processing, self-supervised methods have been developed for phoneme boundary detection and unsupervised speech segmentation. For example, segment-level contrastive predictive coding (SCPC) learns hierarchical representations to identify phoneme-like fragments and boundaries in continuous speech without labels (Bhati et al., 2022). Another work directly formulates unsupervised phoneme boundary detection using contrastive objectives and peak detection (Kreuk et al., 2020). Similar ideas appear in bioacoustic event segmentation, where self-supervised learning is used to detect temporal boundaries in animal vocalizations (Bermant et al., 2022).

In the video domain, self-supervised segmentation has been studied in both spatial and temporal contexts. For example, (Yang et al., 2021b) develops self-supervised video object segmentation method by learning foreground-background separation from object motion cues on unlabeled video streams, and there are some similar subsequent works, e.g., self-supervised deformable attention distillation (Truong et al., 2024), and self-supervised amodal visual object segmentation (Yao et al., 2022). For temporal action segmentation, SSCAP first learns discriminative frame features through self-supervision, and then applies a Co-occurrence Action Parsing algorithm to parse sub-action temporal paths (Wang et al., 2022c). SSTAE proposes a self-supervised temporal autoencoder that performs action segmentation in egocentric videos via reconstruction + clustering (Zhang et al., 2023). More recent work further explores optimal transport / joint alignment + segmentation to achieve fully self-supervised temporal action segmentation, such as VASOT (Ali et al., 2025). These methods share a common characteristic: self-supervised objectives (intra-action consistency, cross-video co-occurrence, prediction/reconstruction objectives, etc) are used to infer temporal structure and boundaries directly from raw sequential data.

Our work brings this segmentation-centric perspective to EEG for the first time. Inspired by advances in speech and video, we design self-supervised objectives tailored to EEG that explicitly target temporal boundary discovery under the assumption that intention formation and cognitive processing are stage-based. Importantly, unlike prior EEG–SSL approaches, our framework does not rely on artificial perturbations or downstream task supervision; instead, segmentation is treated as a structural inference problem driven by endogenous neurodynamic regularities.

## 2.3 Neurodynamic State Discovery and the Limits of Gradient-Based Segmentation

From a cognitive neurodynamics standpoint, the problem of discovering latent stages from EEG is closely related to classical work on brain state segmentation, metastable dynamics, and state-space models of cognition. Probabilistic frameworks such as hidden Markov models (HMMs), switching linear dynamical systems, and related Bayesian change-point models have been widely used to infer latent brain states or transitions from neural time series (Borst & Anderson, 2015; Song et al., 2019; Wang et al., 2022a). Related ideas also appear in EEG microstate analysis, where quasi-stable spatial patterns are identified and interpreted as elementary building blocks of cognition.

While these approaches have provided valuable insights, they typically rely on parametric assumptions, gradient-based inference, or probabilistic state transitions defined over relatively low-dimensional representations. Moreover, they often emphasize population-level structure or short-timescale stability, rather than individualized, long-horizon neurodynamic progression. In practice, the discrete and sparse nature of temporal boundaries makes direct optimization difficult within standard differentiable frameworks.

Similarly, existing self-supervised segmentation methods in speech and video almost exclusively rely on gradient-based optimization, leveraging fully differentiable architectures (CNN / RNN / Transformer / 3D Conv) and losses (contrastive, reconstruction, prediction, co-occurrence, etc.). Segmentation (boundaries/masks) is obtained either by attaching a differentiable head on the network output (e.g., frame-wise logits followed by argmax/CRF), or by simple post-processing such as thresholding or peak picking, with the core parameters still learned by gradients. However, EEG segmentation poses a fundamentally different challenge: the core variable of interest, whether a given time point constitutes a boundary, is discrete, sparse, and non-differentiable. This mismatch motivates the exploration of alternative optimization paradigms. In this work, we adopt an evolutionary search framework that enables direct optimization over candidate segmentations and stage structures. By operating in the discrete space of temporal boundaries, the proposed approach complements probabilistic and gradient-based methods, providing a robust and interpretable mechanism for discovering individualized neurodynamic trajectories.

## 3 Methodology

### 3.1 Experimental Design

Crossing the street is one of the most typical and safety-critical scenarios in human–vehicle interaction (Brill et al., 2024; Fu et al., 2019; Zhou et al., 2025b). When pedestrians cross the road, they are essentially negotiating with traffic/vehicles: Is it safe for me to go now? Misjudgment in this negotiation, whether the pedestrian assumes safety when the vehicle cannot stop in time, or the opposite, can directly lead to traffic accidents for both conventional vehicles and intelligent connected vehicles (Gerogiannis & Bode, 2024). Therefore, the scenario is highly safety critical.

In addition, crossing decisions unfold over time as a cognitive process rather than an instantaneous label (Tian et al., 2024). After perceiving the oncoming flow of vehicles, the brain progresses through several stages: initial perception (seeing the car, estimating distance and speed), risk assessment (will I be hit if I go now?), strategy evaluation (should I wait for this car? the next one? should I walk faster or slower?), and finally, an internal commitment to the decision of crossing or not crossing (Sun et al., 2015; Wang et al., 2025; Zhang et al., 2025b). This process forms a staged neurodynamic trajectory, meaning that crossing decisions are naturally a time-series problem with inherent cognitive boundaries.

From a safety and security perspective, if autonomous or intelligent vehicles can understand a pedestrian's internal decision boundaries, e.g., when they have fully committed to crossing versus when they are merely hesitating or probing, they can: (1) better predict pedestrian behavior and reduce collision risk (Zhou et al., 2024), and (2) enable more fine-grained interaction strategies for V2P/e-HMI systems (De Clercq et al., 2019; Yue et al., 2025). Therefore, in this experimental task, we aim to learn each individual's neurodynamic signatures during safety/danger evaluation and decision commitment, and examine whether these signatures remain stable and distinguishable across different scenarios, thereby providing a foundation for future human–vehicle interaction and operator authentication.

In addition, our experiment includes multiple traffic scenarios (e.g., single versus multiple vehicles, different directions such as one-way or two-way flow). These manipulations systematically alter both risk level and decision difficulty, thereby inducing different cognitive loads and risk assessment patterns in EEG. Based on this, we can test the robustness and generalization of the neurodynamic signatures. If our method is truly learning "an individual's neurodynamic pattern during risk evaluation and decision commitment", then it should: (1) consistently detect similar intention-stage structures across different scenarios; (2) produce boundary patterns and signatures that remain relatively stable for the same person across multiple scenarios; and (3) yield signatures with distinguishable differences across individuals (supporting identity). Therefore, these signatures can serve as real priors for subsequent cybersecure human–vehicle systems.

Specifically, five traffic scenarios were introduced in the experiment, as summarized in Table 1 and illustrated in Figure 1. GIF versions of the stimuli are provided in the supplementary materials. The scenarios were designed in Adobe Animate. The animations were created using Motion Tween (Adobe, 2023) at a frame rate of 24 FPS to produce realistic vehicular motion. Visual cues such as traffic lights, turn signals, and ambient lighting were rendered with keyframe-based flicker effects and glow filters to approximate real-world luminance conditions. Scenarios were presented using PsychoPy v2022.2.5 (Peirce et al., 2019). PsychoPy can simultaneously synchronize EEG signals recording, providing temporal markers for scene onset and recording time stamps for button responses associated with decision making commitment.

In the experiment, each trial began with a 500 ms fixation cross ("+"), displayed at the center of the screen to stabilize visual attention and minimize eye-movement artifacts. Then the animated video clip depicting a road-crossing scene was played. Participants were instructed to imagine themselves as the pedestrian standing at the curb, observing traffic, and to press the "up" key when they judged it was safe to cross. Each animation looped continuously until a response was made, ensuring self-paced decision-making without temporal constraints. Upon response, the next scenario was automatically initiated.

Table 1. Descriptions of the simulated experimental scenarios.

| Traffic Scenarios | Description |
| --- | --- |
| Minimal Traffic Volume | A two-lane road without signalized crosswalks, with no cars approaching within sight. |
| Low Traffic Volume | A two-lane road without signalized crosswalks, with two cars approaching at a slow speed within sight. |
| High Traffic Volume | A two-lane road without signalized crosswalks, with two cars approaching in the adjacent lane and three vehicles approaching in the opposite lane at a moderate speed within sight. |
| Surface-marked Intersection | A four-way intersection with surface-marked crosswalks but without traffic lights. One car is waiting behind the sidewalk on the side opposite to the pedestrian's intended movement, with its right-turn signal flashing and visible within sight. |
| Signalized Intersection | A four-way intersection with surface-marked crosswalks and a traffic light (green/red). One car is waiting behind the sidewalk in the vertical orthogonal direction relative to the pedestrian's intended movement with its right-turn signal flashing. A red traffic light, perpendicular to the pedestrian's movement, is visible at the intersection. |

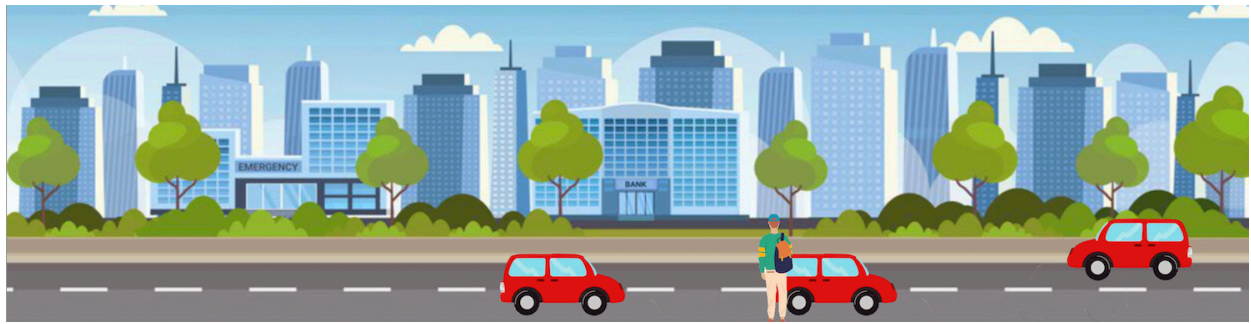

Figure 1. Experimental Stimulus of scenario "High Traffic Volume" (presented as animations to participants).

## 3.2 Data recording

EEG signals was recorded using a 14-channel EPOC X headset, with electrodes positioned according to the international 10–20 system (Klem et al., 1999). Saline-soaked felt pads were used to ensure contact, and electrode impedance was continuously monitored through EmotivPro software (EMOTIV, 2023) until all sensors achieved optimal contact quality. EEG data were sampled at 128 Hz with an online 0.16 Hz first-order high-pass filter applied to remove DC offset.

Twelve subjects (six males and six females, mean age = 24.92 years) participated in the experiment. Informed consent was obtained verbally prior to participation, and the study protocol was approved by the University of Michigan Institutional Review Board (Reference ID: HUM00249262).

EEG signal quality was assessed after importing the data into MNE-Python (Gramfort et al., 2013), which automatically evaluates data completeness and channel integrity; all channels across all participants were therefore retained for analysis. Using the collected EEG data, we applied machine learning methods to predict pedestrians' road-crossing intentions from neural signals, and achieved an AUC of 0.73 (Zhou et al., 2024). This performance exceeds the best AUC reported by recent vision-based approaches that predict pedestrian crossing behavior using pose and positional information (AUC = 0.59) (Zou et al., 2025), providing empirical support for the reliability and informational richness of the EEG recordings.

### 3.3 Preprocessing and Representation

The EEG recordings were preprocessed into bandpower features by Fast Fourier Transform (Cochran et al., 1967), yielding 70 dimensions (14 electrodes × 5 frequency bands). Each trial $j$ is represented as a matrix $X_j \in \mathbb{R}^{T_j \times D}$, where $D = 70$ and $T_j$ is the trial length. Each row of $X_j$ is a feature vector at time $t$:

$$x_t \in \mathbb{R}^D, \quad t = 1, \dots, T_j$$

To remove scale differences across electrodes and bands, each feature dimension is standardized within a trial:

$$x_{t,d} \leftarrow \frac{x_{t,d} - \mu_{j,d}}{\sigma_{j,d} + \epsilon}, \quad d = 1, \dots, D$$

Where $\mu_{j,d}$ and $\sigma_{j,d}$ are the mean and standard deviation of feature $d$ in trial $j$, and $\epsilon = 10^{-6}$ prevents division by zero.

Each candidate $c$ consists of three ordered change-points as stage boundaries per trial:

$$\tau^j = \left(t_1^j, t_2^j, t_3^j\right), \quad 0 < t_1^j < t_2^j < t_3^j < T_j$$

which partition the trial into four contiguous stages, and stage-specific weights:

$$w_k \in \mathbb{R}^D, \quad \|w_k\|_0 \leq K_s \text{ (sparse)}$$

where $K_s$ is the maximum umber of nonzero entries allowed per stage. To promote interpretability and physiological plausibility, we set $K_s$ to 5-7, corresponding to the most informative electrode-band combinations.

within each stage $k$, the EEG is projected into a one-dimensional latent trajectory by:

$$z_k(t) = w_k^\top x_t$$

which makes subsequent dynamics and distribution shifts tractable.

### 3.4 Objective Function

We design a multi-term objective (including temporal predictability, inter-stage distinctness, cross-trial consistency, and parsimony) to evaluate the quality of a candidate segmentation. Each term is chosen to reflect a basic cognitive-neurodynamic principle: cognitive states tend to be locally stable, transitions reflect qualitative shifts in processing, and the stage sequence of a repeated task should be reproducible across trials.

1. Predictability within stages (stage stability). For each projected trajectory $z_k$, we fit an autoregressive model of order 1 (AR(1)):

$$z_t \approx a z_{t-1} + b + \epsilon_t$$

and compute the mean squared residual error. This terms rewards segmentations that produce stages with internally predictable dynamics, aligning with the neurodynamic intuition that a metastable cognitive state should exhibit relatively structured temporal evolution, whereas prediction error should increase near transitions where the underlying processing regime changes. This formulation also aligns with classical time-series segmentation assumptions, where each segment is modeled as an approximately stationary process and fitted using autoregressive or state-space models, as seen in HMMs and HSMMs. In cognitive neurodynamics, the same idea appears in "brain states" concepts, where within state spectral or connectivity patterns are expected to remain relatively stable over short intervals.

2. Boundary contrast (regime shift between processing stages). To ensure adjacent stages correspond to genuinely different cognitive states rather than smooth fluctuations, we compute the symmetric Kullback–Leibler (KL) divergence between Gaussian fits of neighboring stage distributions. Let $\mu_1, \sigma_1^2$ and $\mu_2, \sigma_2^2$ denote the means and variances of two consecutive stages, the symmetric divergence is:

$$D_{symKL} = \frac{1}{2}\left(\frac{\sigma_1^2}{\sigma_2^2} + \frac{\sigma_2^2}{\sigma_1^2} + (\mu_1 - \mu_2)^2 \left(\frac{1}{\sigma_1^2} + \frac{1}{\sigma_2^2}\right) - 2\right)$$

This term is maximized because if a time point is designated as a boundary, the neural activity patterns on either side of that boundary should exhibit a pronounced change, reflecting a rapid transition between qualitatively different processing stages (e.g., from perceptual assessment to risk evaluation or decision commitment), rather than minor drift.

3. Cross-trial alignment (reproducible stage sequence). Because trials are repetitions of the same task, transitions should occur at consistent relative positions across trials. We therefore penalize the squared deviation between each trial's normalized boundary positions $\alpha^j = \tau^j/T_j$ and the across-trial mean $\bar{\alpha}$. This encourages the model to discover reproducible stage structures rather than overfitting to idiosyncratic noise in individual trials. Although response durations may vary across trials, the internal cognitive process underlying a "safe-to-cross" decision generally follows an invariant sequence of stages—perception, risk assessment, deliberation, and decision commitment. Accordingly, the boundaries we aim to identify correspond to recurring neurodynamic transitions that can be temporally aligned across repetitions. This perspective is consistent with self-supervised segmentation in video and speech, where cross-sample co-occurrence and alignment are leveraged to identify recurring action segments or phoneme-like units.

4. Sparsity (interpretable stage-wise signatures). To improve interpretability and isolate compact neurodynamic signatures, we include an L1 penalty $\lambda_1 \|w_k\|_1$ that drives most feature weights to zero, leaving only the few most informative electrode-band pairs for each stage. This supports stage-wise attribution by encouraging each discovered cognitive state to be characterized by a sparse set of salient neurophysiological features, rather than diffuse patterns that are difficult to interpret and less robust to noise.

The final objective for a candidate $c$ is:

$$\mathcal{L}(c) = \sum_{j,k}(L_{AR1}(z_{jk})) - \lambda_{bdry}\sum_{j,k}(D_{symKL}(z_{jk}, z_{j,k+1}) + \lambda_{align}\sum_{j}\|\alpha^j - \bar{\alpha}\|^2 + \lambda_1\sum_{k}\|w_k\|_1$$

In our implementation, we set $\lambda_{bdry} = 0.3$, $\lambda_{align} = 0.1$, and $\lambda_1 = 0.05$, values tuned to balance discriminability and stability. These values were chosen empirically after preliminary runs: higher boundary weights caused over-segmentation, while stronger sparsity could overly degrade predictability.

### 3.5 Evolutionary Optimization

In current research works that integrate evolutional computation (EC) with SSL, there are two dominant paradigms (Vinhas et al., 2025). The first uses EC to support SSL by optimizing pseudo-labels, data transformations, architectures, or learning rules in pretext tasks and downstream fine-tuning, while the second uses SSL to support EC by providing learned representations, smarter variation operators, or surrogate fitness functions. Our work belongs nominally to the former category (EC for SSL), but departs fundamentally in what is being optimized and why.

Most existing EC-for-SSL approaches treat evolution as a mechanism for configuration optimization, which can be broadly categorized as follows.

Pretext task – Dataset (evolving "data" or pseudo-labels). This line of work uses EC to adjust the data or pseudo-labels used in self-supervised tasks. For example, GenNAS uses evolutionary algorithms to search for synthetic signals or pseudo-label combinations that yield better feature representations during the pretext stage (Li et al., 2021). Another strategy evolves input transformations or augmentation functions, such as cropping, noise injection, masking, or combinations thereof, in order to automatically identify effective augmentation for contrastive or reconstruction tasks. These methods are most commonly applied to image representation learning.

Pretext task – Topology (evolving network architecture). The second category uses EC to evolve encoder or autoencoder architecture, including convolutional layer depth, channel width, skip connections, or whether to include a projector head, etc. Most studies focus on autoencoder / reconstruction-based SSL, such as EvoAE (Lander & Shang, 2015), EvoVAE (Chen et al., 2020), and EUDNN (Sun et al., 2018). These approaches are primarily used for self-supervised pretraining of image or text features, followed by classification or retrieval as downstream tasks.

Pretext task – Learning (evolving learning rules and hyperparameters). A third group evolves how the model learns, including learning rates, optimizer configurations, loss weights, or training schedules. In some cases, the self-supervised objective is also included in the search space. However, most of these still lean toward reconstruction loss and architecture search, and very few addresses inherently non-differentiable objectives such as sparse temporal boundary selection.

Downstream stage (evolutionary fine-tuning and model selection). A small portion of work applies EC during the downstream fine-tuning stage to select the best encoder, task-specific head or hyperparameters based on limited labeled data. The performance is typically assessed using downstream metrics such as accuracy, most commonly in speech recognition, text classification, or image classification tasks.

Despite their diversity, evolution is used to optimize the configuration in SSL, rather than the self-supervised inference target itself. Evolution operates on how data are transformed, how models are structured, or how learning proceeds, while the core learning problem, typically continuous, differentiable representation learning, is still solved by gradient-based optimization. Our approach, however, uses EC to directly optimize the structural outcome of self-supervised learning, namely sparse temporal boundary selection and segmentation. Rather than evolving architectures, hyperparameters, or pretext-task definitions, we evolve the segmentation solution itself.

Specifically, we design an evolutionary algorithm to search the space of possible temporal segmentations that jointly refines temporal boundaries and stage-specific feature weights. The procedure is as follows:

1. Initialization.
   Each candidate $c$ is initialized with sparse weights $w_k \in \mathbb{R}^D$ and change points $\tau^j = (t_1^j, t_2^j, t_3^j)$ positioned near quartiles with added random jitter.
2. Population loop
   For each generation $g = 1, \ldots G$ with population size $P = 60$:
   1) Evaluation: compute the objective $\mathcal{L}(c)$ for each candidate.
   2) Selection: retain the top $E = 6$ elites unchanged. The remaining $P - E$ offspring are generated by rank-proportional selection.

3) Crossover: given two parents $c^{(1)}, c^{(2)}$, offspring boundaries are set to the elementwise median:

$$\tau_{child} = median(\tau^{(1)}, \tau^{(2)})$$

and weights are averaged,

$$w_{k,child} = \frac{1}{2}(w_k^{(1)} + w_k^{(2)})$$

4) Mutation: with probability $p_\tau = 0.1$, each boundary is jittered by up to $\pm 0.05 T_j$. With probability $p_w = 0.1$, each weight entry is perturbed by Gussian noise,

$$w_{k,d} \leftarrow w_{k,d} + \epsilon, \quad \epsilon \sim \mathcal{N}(0, 0.1^2)$$

Or a nonzero feature index is swapped with a zero feature to maintain sparsity.
3. Early stopping
If the best objective value does not improve for 15 generations, evolution stops. In practice, convergence is typically achieved within 50 generations, yielding stable segmentations and interpretable feature masks.

### 3.6 Baseline Methods

### 3.6.1 Baseline Selection Rationale

To benchmark our self-supervised evolutionary learning, we implemented three strong unsupervised segmentation baselines. All methods enforce a minimum segment length of 8% of the trial.

In the statistical change-point detection category, segmentation is treated as the task of "finding time points where the statistical properties of a time series change". In general time-series research, comprehensive surveys (Aminikhanghahi & Cook, 2017; Truong et al., 2020) classify these methods into probabilistic models, cost-function + dynamic programming approaches, and kernel-based approaches. Representative methods include:

**Bayesian Online Change-Point Detection (BOCPD).** Proposed by (Adams & MacKay, 2007) for online Bayesian change-point detection, explicitly modeling the posterior distribution of the "run length" and updating the probability of a new segment at every new sample. Widely used for online state change detection in medical and physiological time series (e.g., heart rate, respiration, and other biosignals) (Gee et al., 2018).

**Kernel Methods (Kernel Change-Point Detection, KCPD).** Use kernel techniques to map high-dimensional observations into a feature space where distribution shifts can be detected using criteria such as Maximum Mean Discrepancy (MMD) (Gretton et al., 2012). (Arlot et al., 2019) provided foundational theory and algorithms; both Truong's survey and the ruptures library treat kernel CPD as a major category (Truong et al., 2018, 2020). Suitable for high-dimensional, nonlinear time series, aligning well with high-dimensional EEG settings.

**PELT (Pruned Exact Linear Time) and Other Penalized Cost + DP Methods.** PELT is a typical "cost function + penalty + dynamic programming" framework (Killick et al., 2012). Truong's survey treats it as a strong baseline for offline multiple change-point detection, with near-linear computational complexity (Truong et al., 2020).

These methods were originally mainstream tools for general time-series segmentation and are often adapted for EEG (especially for sleep stage transitions, brain state switching, noise segment detection, etc.) (Sinn et al., 2012a), but they are fundamentally modality-agnostic. We therefore select these three classic, broadly reproducible, and widely recognized methods as our baselines for comparison.

In addition, there are another two categories of methods for segmentation on time-series data. One is model-driven state-switching methods, such as Hidden Markov Models (HMM)/ Hidden Semi-Markov Models (HSMM), and switching state-space models (Borst & Anderson, 2015; Song et al., 2019; Wang et al., 2022a). The model-driven approach explicitly models latent states and their transition probabilities, making it well suited for interpreting sequences of putative "brain states" (Williams et al., 2018). However, they typically require pre-specification of the number of states (Sgouralis & Pressé, 2017), and rely on EM/Baum–Welch–based training procedures, which become computationally heavy for high-dimensional EEG signals and long temporal sequences (Siddiqi & Moore, 2005). The other is hybrid deep learning + encoding–segmentation methods, such as autoencoder + HMM, VAE + clustering, etc., commonly used in epilepsy or sleep studies (Rivas-Carrillo et al., 2023; Yang et al., 2021a; Yıldız et al., 2022). Although these approaches incorporate representation learning, they are generally designed to support specific downstream tasks such as event or episode detection, with performance evaluated primarily using task-level accuracy or classification metrics rather than boundary detection quality itself (Yıldız et al., 2022; Zhao et al., 2024b).

Because our work treats temporal boundary discovery as the primary inference objective, rather than as an intermediate step toward downstream classification, and avoids assumptions about the number of latent states or task-specific labels, the problem setting is fundamentally different from those addressed by these methods. Therefore, they are not included as baseline comparisons in our evaluation.

### 3.6.2 Baseline implementation

**(1) PELT with L2 cost**

Pruned Exact Linear Time (PELT) algorithm is a dynamic programming approach for change-point detection (Killick et al., 2012). It is a computationally efficient ($O(T)$) and well-suited for multivariate Gaussian-like signals (Maidstone et al., 2017).

For each trial $X \in \mathbb{R}^{T \times D}$, PELT minimizes the penalized residual sum of squares:

$$\min_{\tau_1, \tau_2, \tau_3} \sum_{k=0}^{3} \sum_{t=\tau_k+1}^{\tau_{k+1}} \|x_t - \bar{x}_k\|_2^2 + \beta \cdot m$$

where $\bar{x}_k$ is the segment mean, $m$ the number of change-points, and $\beta$ a penalty controlling over-segmentation.

For the output, three boundaries closets to quartiles, repaired to meet minimum length, are retained.

### (2) Kernel Change-Point Detection (RBF kernel)

We use the KernalCPD method with an RBF kernel to capture non-linear distributional changes across EEG features (Harchaoui et al., 2008; Sinn et al., 2012b).

For trial $X$, kernelCPD minimizes within-segment kernel variance:

$$\min_{\tau_1, \tau_2, \tau_3} \sum_{k=0}^{3} \sum_{t=\tau_k+1}^{\tau_{k+1}} K(x_t, x_t) - \frac{1}{n_k} \sum_{s,t \in S_k} K(x_s, x_t)$$

where $K(x_s, x_t) = \exp(-\|x_s - x_t\|^2 / 2\sigma^2)$ and $n_k$ is the segment size.

Boundaries are chosen to maximize kernel contrast, then enforced to quartilelike positions if underdetermined.

### (3) Bayesian Online Change-Point Detection (BOCPD) with PCA(1)

BOCPD provides an online Bayesian approach with well-calibrated uncertainty to infer the timing of cognitive state transitions (Adams & MacKay, 2007; Koerner et al., 2017). The algorithmic details are described as follows. For projection:

$$z_t = u_1^\top (x_t - \bar{x}), \quad u_1 = arg \max_{u: \|u\|=1} \|X_u\|_2$$

Run-length $r_t$ follows a constant hazard $H = 1/\lambda$ with $\lambda = 200$. Predictive distribution for new points is Student-t:

$$p(z_t | z_{1:t-1}, r_{t-1}) = t_\nu(\mu, \sigma^2)$$

with parameters updated recursively from sufficient statistics.

Change-points are identified at local minima in run-length with high negative log predictive scores. The top three are selected subject to minimum length.

### 3.7 Evaluation Metrics

Because EEG segmentation is unsupervised, we cannot rely on ground-truth labels to assess accuracy. Instead, we design a suite of intrinsic and stability-based metrics that capture how well a segmentation explains the signal, how distinct adjacent stages are, and how reproducible results are across top candidates.

### 3.7.1 Intrinsic Predictability Gain

As computed in Section 3.2, we further computed the mean squared error (MSE) of the AR(1) fit within each stage, $L_{stage}$, and compare it against the MSE of a global AR(1) model fit to the entire trial, $L_{global}$. The AR predictability gain is defined as:

$$AR - Gain = \max\left(0, \frac{L_{global} - \overline{L_{stage}}}{|L_{global}| + \epsilon}\right)$$

Stable cognitive states should be internally predictable. Segmentations that isolate such states thus achieve lower AR(1) error compared to modeling the whole trial as one process.

### 3.7.2 Out-of-Segment Generalization

We test whether stage-specific AR models generalize beyond their training portion. For each stage segment of length $n$, we fit an AR(1) model on the first half and evaluate on the second half, yielding a test error $L_{test}$. If segmentation captures meaningful states, then models trained on one part of a stage should predict the rest, thus demonstrating temporal coherence rather than overfitting.

The generalization gain is defined as:

$$Gen - Gain = \max\left(0, \frac{L_{global} - \overline{L_{test}}}{|L_{global}| + \epsilon}\right)$$

### 3.7.3 Boundary Contrast (Separation)

To measure distinctness between adjacent stages, we compare their distributions using a symmetric KL divergence (see Section 3.2). The average across three boundaries gives the boundary contrast score.

### 3.7.4 Stability Across Top-K Candidates

Because SSEL saves the best $K = 5$ candidates, we assess stability in two ways:

**Boundary stability**: For each trial and boundary, we compute the normalized mean absolute deviation (MAD) of change-points across candidates:

$$Stability = \frac{MAD(\{\tau^{(1)}, \dots, \tau^{(K)}\})}{T}$$

Lower values indicate consistent boundary placement.

**Channel-set stability**: For each stage, we take the top-8 features (by absolute weight) from each candidate and compute the pairwise Jaccard index:

$$J(A, B) = \frac{|A \cap B|}{|A \cup B|}$$

The average across all candidate pairs quantifies how consistently the algorithm selects the same features. Since robust methods should yield similar segmentations and channel selections across near-optimal runs, higher values are better.

## 4 Results

This section aims to highlight the dual interpretability of our approach: on one hand, the identified stages and associated feature maps reveal neurophysiologically plausible channel–band signatures across cognitive stages, demonstrating the method's ability to capture meaningful brain dynamics; on the other hand, the evolutionary trajectories provide algorithmic transparency, showing how candidate solutions progressively refine through exploration and exploitation within a structured fitness landscape.

### 4.1 Segmentation Performance Against Baselines

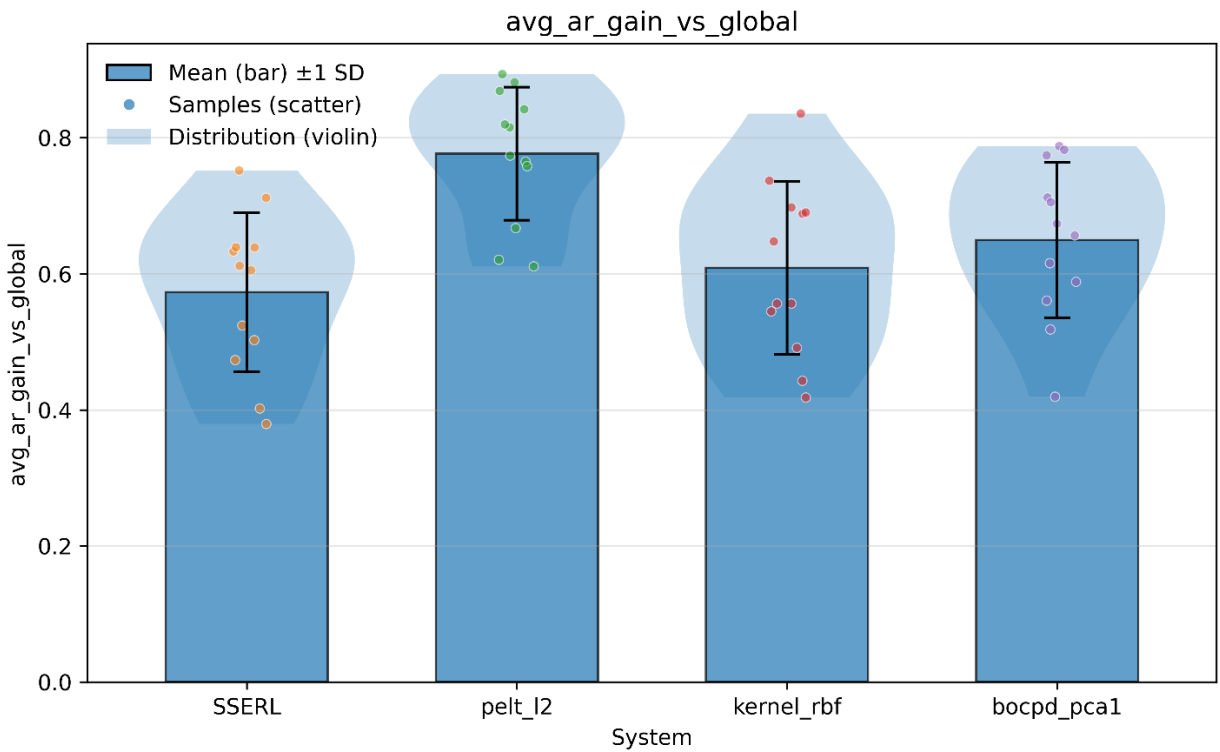

Figure 1. Intrinsic Predictability Gain (AR–Gain) across methods.

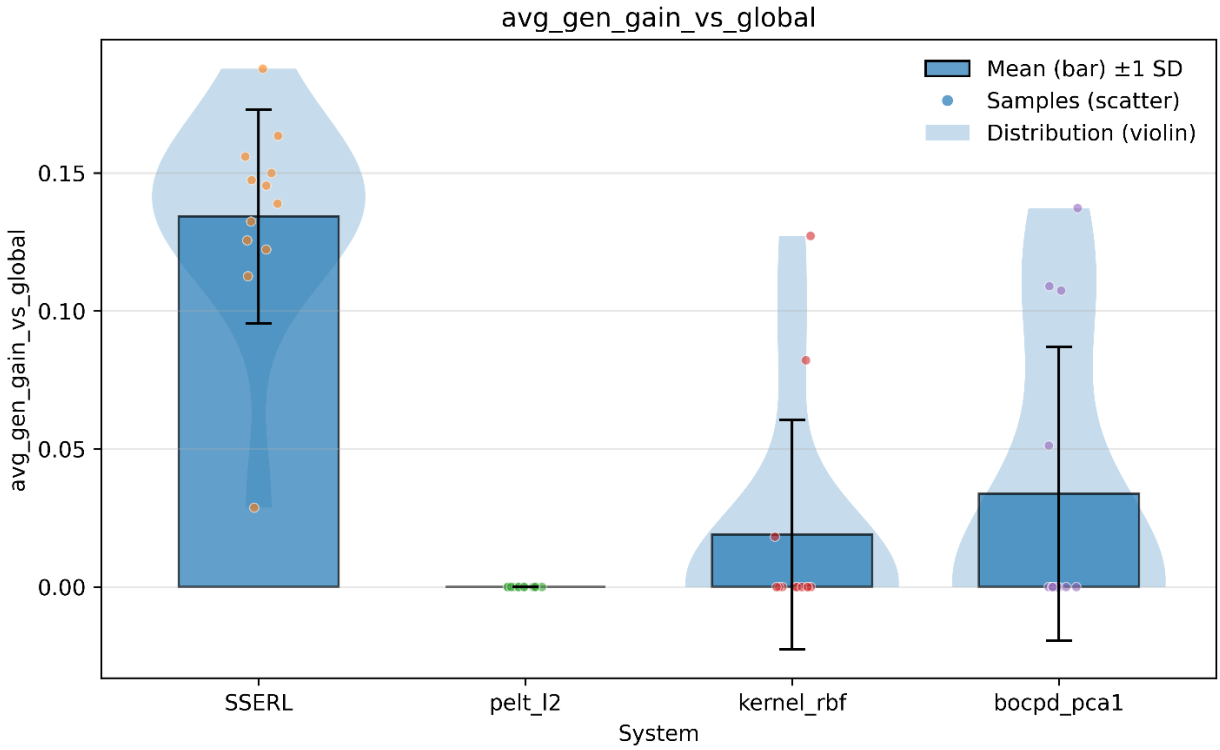

Figure 2. Out-of-Segment Generalization Gain (Gen–Gain) across methods.

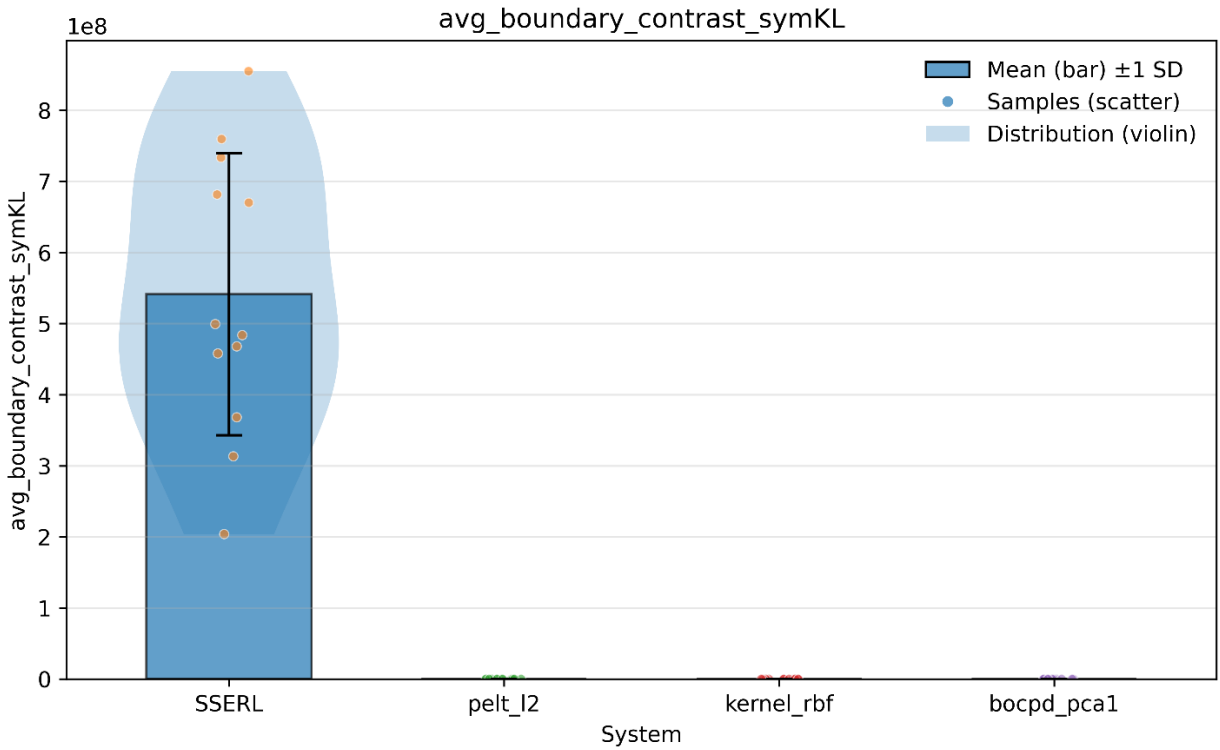

Figure 3. Boundary Contrast across methods.

To assess whether segmentation methods differed significantly across the three evaluation metrics, we employed non-parametric, within-subject analyses. The Friedman test was used to evaluate overall differences, complemented by Kendall's W to estimate effect size (interpreted as weak: 0.1–0.3, moderate: 0.3–0.5, strong: >0.5). Post-hoc comparisons were conducted using Wilcoxon signed-rank tests with Holm–Bonferroni correction for multiple comparisons. For each pairwise comparison, we additionally report the standardized effect size $r = |z|/\sqrt{n}$.

### 4.1.1 Intrinsic Predictability Gain (AR–Gain)

The Friedman test indicated a significant effect of method on AR–Gain, χ2(3)=15.40, p=0.0015, with a moderate effect size (Kendall's W=0.43, n=12). Post-hoc comparisons revealed that pelt_l2 achieved significantly higher AR–Gain than both SSEL (z=−2.98, pHolm=0.0059, r=0.86) and kernel_rbf (z=−2.98, pHolm=0.0059, r=0.86). A trend favoring pelt_l2 over bocpd_pca1 (pHolm=0.084) was also observed, but did not survive correction. Other pairwise contrasts were not significant. These results suggest that pelt_l2 provides the strongest within-stage predictability, while SSEL performs comparatively weaker on this metric.

From a neurodynamic perspective, this outcome suggests that pelt_l2 favors segmentations that impose stronger local stationarity, effectively prioritizing internally homogeneous states. By contrast, SSEL permits greater within-stage variability, which may reflect the fact that real cognitive stages, particularly deliberative or evaluative phases, are not strictly stationary but exhibit structured fluctuations as evidence accumulates and uncertainty is resolved. Thus, slightly reduced AR–Gain does not necessarily indicate inferior cognitive modeling, but rather a relaxation of overly rigid stability assumptions in favor of richer neurodynamic trajectories.

### 4.1.2 Out-of-Segment Generalization Gain (Gen–Gain)

For Gen–Gain, the Friedman test revealed a highly significant method effect, χ2(3)=27.26, p<5.2×10−6, with a strong effect size (Kendall's W=0.76, n=12). Post-hoc analyses indicated that SSEL performed significantly better than each of the three baselines. SSEL's Gen–Gain was significantly higher than kernel_rbf (z=−3.06, pHolm=0.0029, r=0.88), pelt_l2 (z=−3.06, pHolm=0.0029, r=0.88), and bocpd_pca1 (z=−2.98, pHolm=0.0039, r=0.86), with very large effect sizes. No significant differences emerged among the three baseline methods. These findings demonstrate that SSEL generalized substantially better to capture temporal coherence across stage halves.

This strong generalization indicates that SSEL identifies stable cognitive stages whose internal dynamics are consistent across time, even when evaluated on held-out portions of the same stage. This suggests that the discovered segments correspond to genuine cognitive regimes, such as sustained evidence evaluation or motor readiness, rather than transient patterns tied to local noise or momentary fluctuations. Importantly, it implies that these stages reflect enduring internal processing modes, not arbitrary temporal partitions.

### 4.1.3 Boundary Contrast (symmetric KL divergence)

A significant effect of method was also observed for Boundary Contrast, $\chi2(3)=24.40$, $p=2.06\times10^{-5}$, with a strong effect size (Kendall's $W=0.68$, $n=12$). Post-hoc tests showed that SSEL exhibited significantly higher boundary separation than all three baselines: pelt_l2, kernel_rbf, and bocpd_pca1 all yielded $z=-3.06$, $pHolm=0.0029$, $r=0.88$. Pairwise comparisons among the baseline methods did not reach statistical significance following correction. These results indicate that SSEL is substantially more effective at delineating distinct state boundaries compared to the other methods.

Higher boundary contrast implies that transitions identified by SSEL correspond to abrupt shifts between qualitatively different processing stages, rather than gradual drift. This is consistent with models of decision making in which commitment emerges through relatively sharp transitions, such as the shift from deliberation to action, despite variability in overall response time. The results therefore support the interpretation of SSEL boundaries as meaningful cognitive state transitions.

### 4.1.4 Trial Segmentation Visualization

To illustrate how the different segmentation methods operate on raw EEG-derived features, we present segmentation results from a representative subject across five trials (see Figures 4&5). Segmentation was applied to the projected trial trajectories (z-values along samples), with vertical dashed lines indicating the detected change-points.

Among the four methods, bocpd_pca1 yielded stage demarcations that were overly consistent, with little variability across runs. While such stability may appear desirable, it is implausible in the context of our trial design, where diversity in scenario complexity should naturally introduce variability in boundary placement. By contrast, our SSEL method more faithfully followed natural oscillatory transitions and demonstrated stronger alignment with both behavioral observations and retrospective analyses. For example, in Trial 1, the scenario was very simple: once the subject perceived the environment, they quickly transitioned to action initiation, resulting in very short intermediate stages (Stages 2 and 3). A similar pattern was seen in Trial 2, though the introduction of additional elements in the scene prolonged the perceptual stage. In Trial 3, with multiple traffic flows creating a difficult crossing decision, SSEL revealed a clear stage corresponding to evidence accumulation before decision commitment; even after perceptual clarity was achieved, the subject withheld action until the right opportunity, reflected in a sustained Stage 3 of motor readiness.

These examples illustrate that SSEL adapts boundary placement to task difficulty and internal deliberation demands, revealing extended intermediate stages when decisions require prolonged evidence accumulation, and compressed stages when the perceptual–action mapping is straightforward. This flexibility is consistent with cognitive models in which stage order is preserved but stage duration varies with uncertainty and environmental complexity.

For kernel_rbf and pelt_l2, both introduced greater variability in boundary placement compared to bocpd_pca1. Notably, pelt_l2 produced finer-grained segmentations, likely benefiting from its globally optimal dynamic programming formulation. This granularity is consistent with the superior performance metrics reported earlier, yet the method still showed limited plausibility in aligning boundaries with meaningful state transitions. For example, in Trials 1 and 2, some identified boundaries appeared disconnected from the natural perceptual–action progression observed in the behavioral data.

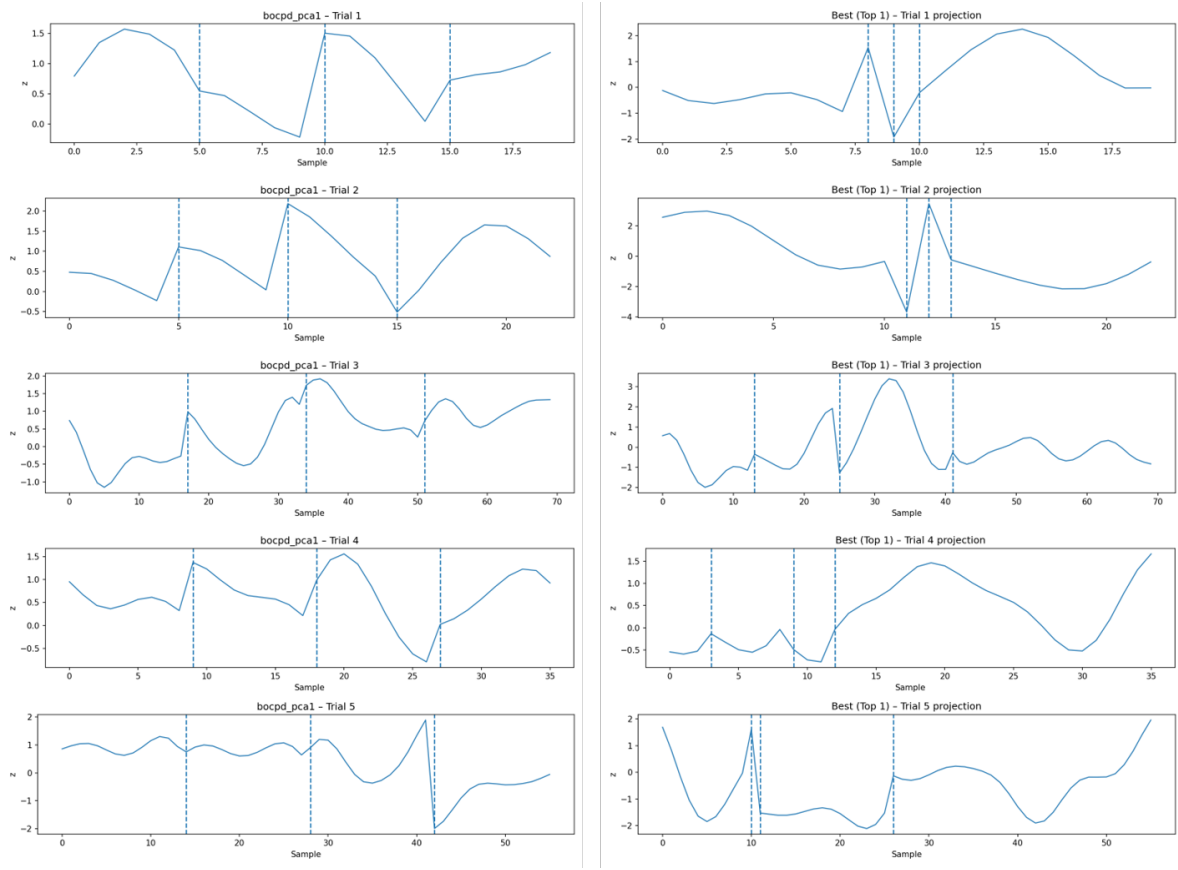

Figure 4. Segmentation of five trials from a representative subject, with bocpd_pca1 (left) and SSEL (right).

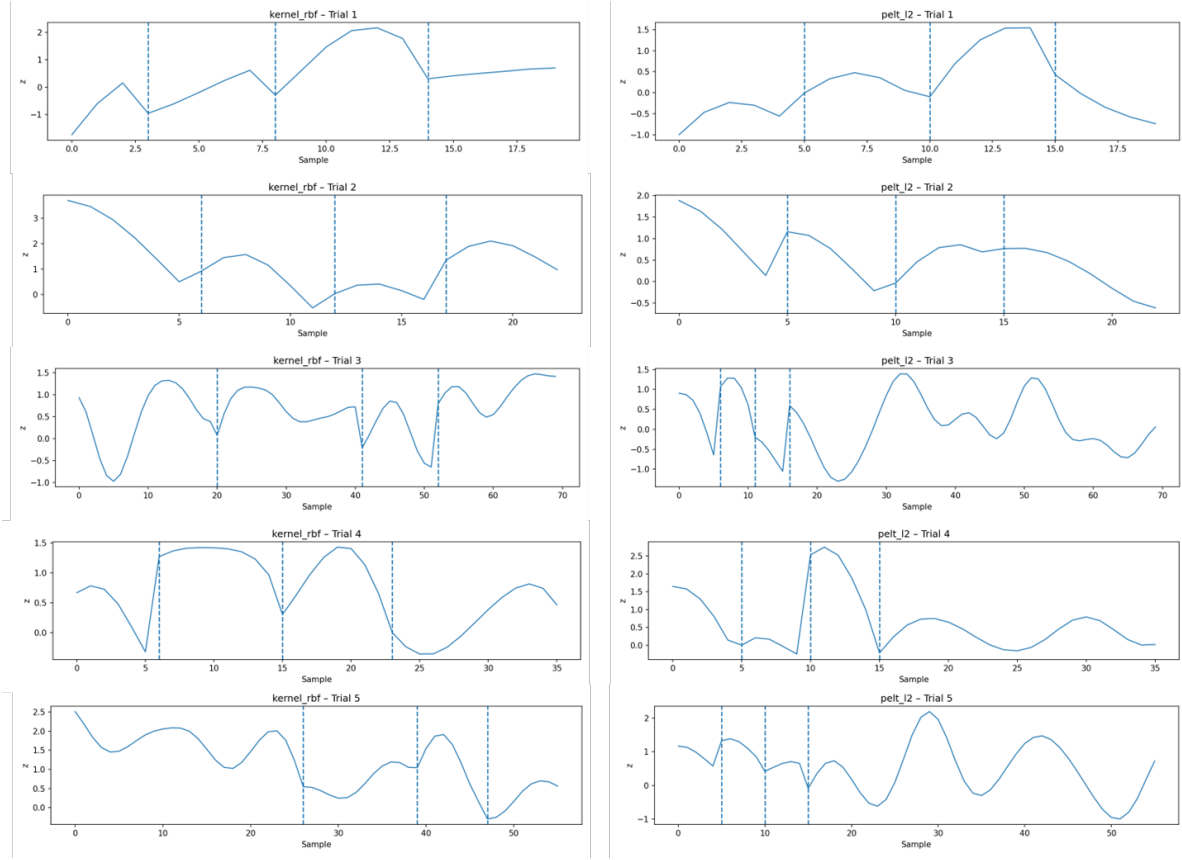

Figure 5. Segmentation of five trials from a representative subject, with kernel_rbf (left) and pelt_l2 (right)

**4.2 Identified Feature Plausibility**

To further investigate the basis of segmentation, we aggregated the top-channel weights identified across all subjects and trials. The feature maps were constructed by summing the normalized weights of selected features (band × channel pairs) for each of the four stages. This produced two-dimensional heatmaps, where the y-axis represents five frequency bands (Theta, Alpha, BetaL, BetaH, Gamma) and the x-axis corresponds to the 14 EEG channels in the Emotiv Epoch X headset. Color intensity reflects the aggregated weight, thus capturing the relative importance of features across the population.

The maps revealed distinct stage-specific signatures. In Stage 1, strong positive weights were observed for frontal and parietal sites (e.g., F7, F8, P7, FC6) in the Theta and Gamma bands, indicating early engagement of frontal oscillations. This pattern is highly consistent with prior neurophysiological findings reported in the original dataset paper (Zhou et al., 2024), which emphasize frontal–parietal synchrony during perceptual onset.

In Stage 2, weights redistributed, with BetaL activity at FC5 and parietal sites (P7, P8) emerging as salient, alongside Alpha modulation at FC5. The presence of BetaL at this stage is consistent with its role in top–down attentional control, reflecting the suppression of irrelevant sensory information and the stabilization of evidence accumulation (Dubey et al., 2023). Concurrent Alpha activity suggests selective inhibition of distracting inputs (Foxe & Snyder, 2011), supporting the transition from perception to evaluation.

In Stage 3, Theta power at F7 became dominant, while BetaH at F8 was markedly suppressed. This combination reflects a shift toward motor readiness: frontal Theta is associated with cognitive control and sustained evidence monitoring (Karakaş, 2020), whereas BetaH suppression in frontal and motor-related regions is widely linked to the release of motor inhibition in preparation for action execution (Wagner et al., 2017). Thus, the Stage 3 map plausibly captures the neural substrate of decision commitment and anticipatory motor control.

Stage 4 demonstrated the strongest BetaL weights at FC5, a pattern well aligned with the literature on movement-related event-related desynchronization (ERD) (Peter et al., 2022). Right-hand motor execution is typically accompanied by bilateral BetaL ERD, especially over central and frontal electrodes. The persistence of this signature at the final stage provides strong convergent evidence that the SSEL framework not only delineates stages algorithmically but also recovers neurophysiologically interpretable features associated with motor output.

Taken together, the orderly evolution of channel–band signatures across stages supports a trajectory-based view of cognition, in which perceptual encoding, evaluative control, motor preparation, and execution emerge as temporally organized neurodynamic stages rather than isolated neural markers. This convergence between algorithmic segmentation and established neurophysiology provides strong evidence that the discovered stages correspond to meaningful cognitive processes.

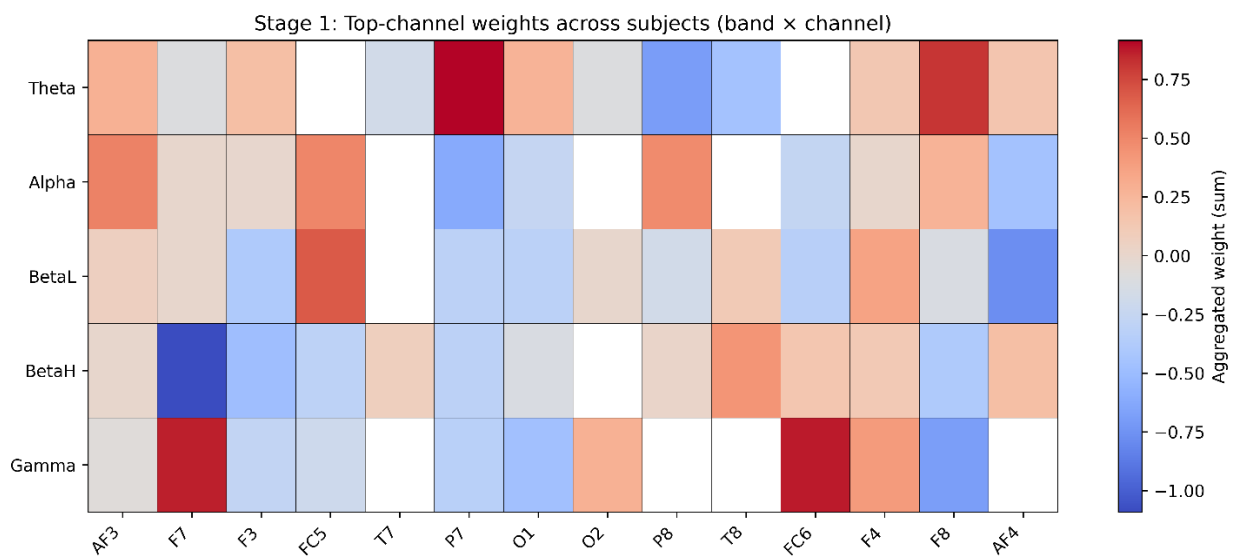

Figure 6. Stage 1 feature contribution heatmap showing aggregated top-channel weights (band × channel).

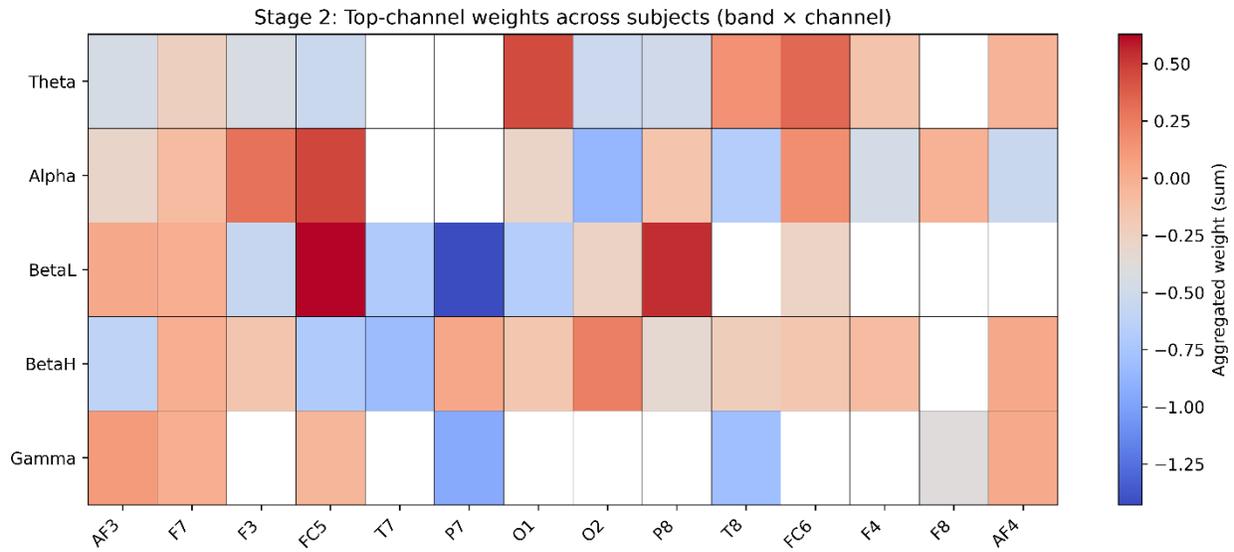

Figure 7. Stage 2 feature contribution heatmap showing aggregated top-channel weights (band × channel).

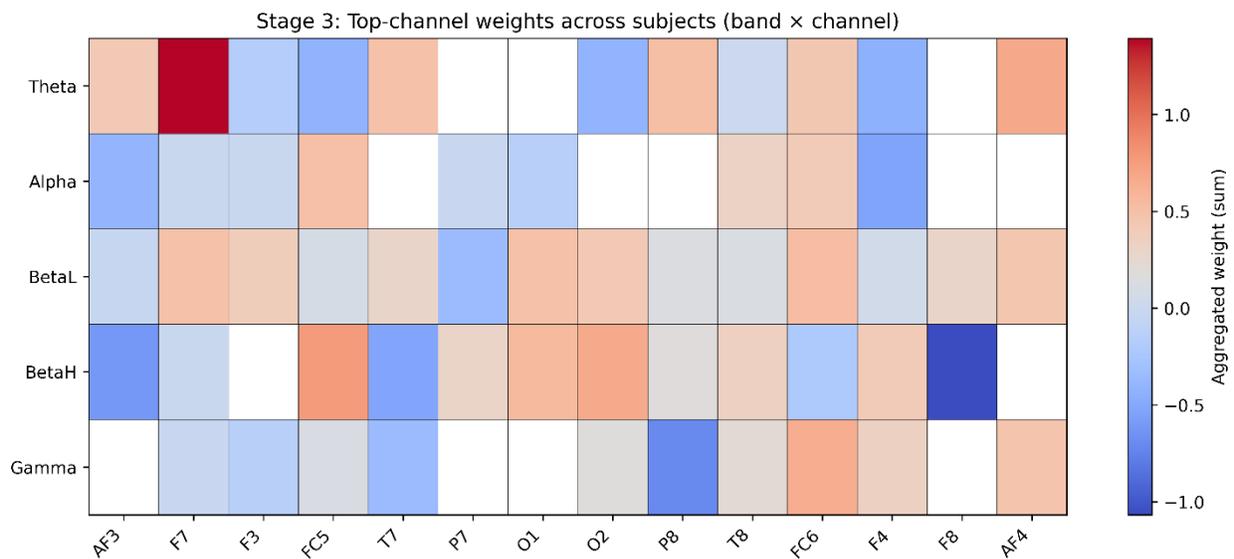

Figure 8. Stage 3 feature contribution heatmap showing aggregated top-channel weights (band × channel)

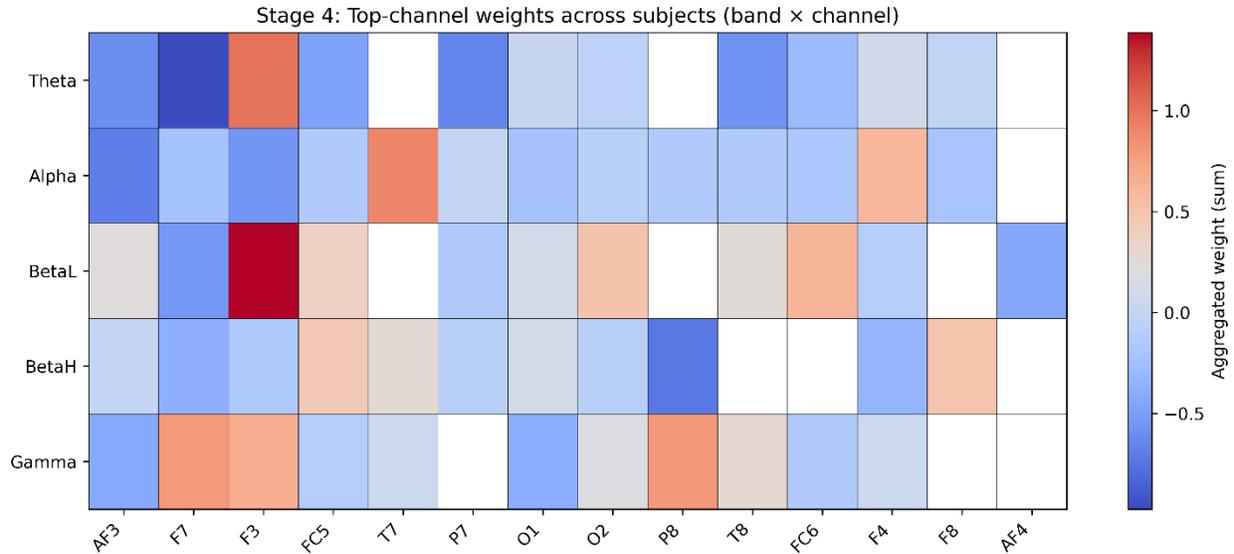

Figure 9. Stage 4 feature contribution heatmap showing aggregated top-channel weights (band × channel).

### 4.3 Evolutionary Search Landscape

To visualize the evolutionary optimization process, we plotted snapshots of candidate solutions projected into a two-dimensional principal component space at generations 30, 60, and 90. The background colormap represents the interpolated fitness landscape (lower values indicating better scores), and candidates are displayed as scattered points with a separate colormap indicating their fitness. This projection was obtained by applying PCA to candidate vectors across generations, with scores interpolated using cubic grid fitting.

At generation 30, candidates were dispersed broadly across the space, reflecting extensive exploration. By generation 60, the distribution had begun converging toward the valley floor, with clusters forming in lower-scoring regions. At generation 90, the candidate distribution was much more concentrated, with most points localized in a narrow band along the global minima. These progressive changes demonstrate the balance of exploration and exploitation in the evolutionary search, with early stages sampling broadly and later stages focusing on refinement around promising regions.

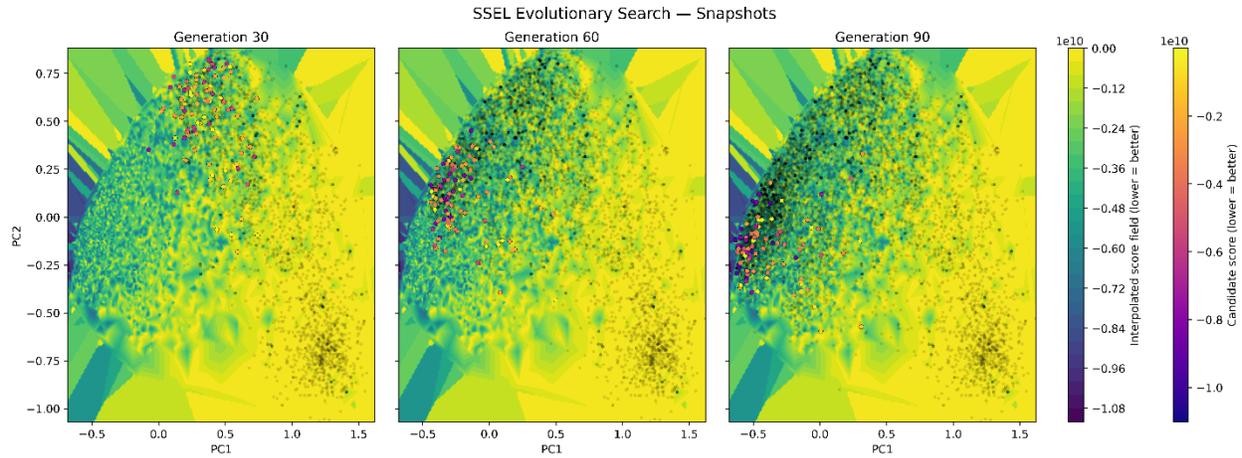

Figure 10. Snapshots of the evolutionary search process at generations 30, 60, and 90. Candidates are plotted in PCA space, with color intensity indicating fitness (lower = better). The background field represents the interpolated fitness landscape.

To better capture the dynamics of the evolutionary search, we constructed 3D surface plots of the interpolated fitness landscape with the trajectory of the best candidate per generation overlaid (Figures 11-13). Extreme outliers were clipped to stabilize the colormap. Best candidates were projected into PCA space and connected temporally, with generation encoded by color along the trajectory. To enhance interpretability, each 3D surface plot was paired with a top-view projection, where the former emphasizes surface morphology and the latter reveals the optimization path across the valley floor.

For both Subject A and Subject B, the trajectory showed an early descent into the fitness valley followed by gradual fine-tuning along the valley ridge. The temporal color progression confirmed a consistent improvement over generations. When the crossover rate was reduced from 0.7 to 0.2, as shown in Figure 12 vs Figure 13, the trajectory became less direct and exhibited detours in PCA space, suggesting the effect of crossover on recombination efficiency and convergence speed.

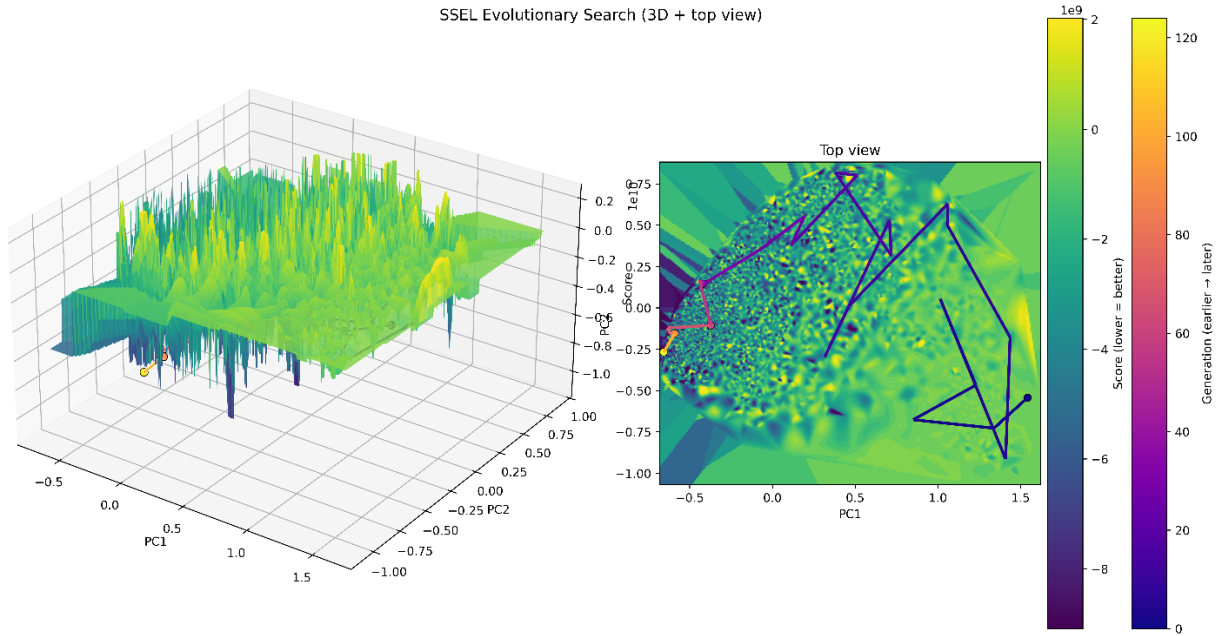

Figure 11. Joint 3D and top-view plots of the evolutionary search landscape for Subject A, showing the fitness surface (colored by score), the trajectory of the best candidate per generation (colored by temporal order), and the alignment of the trajectory with the valley floor.

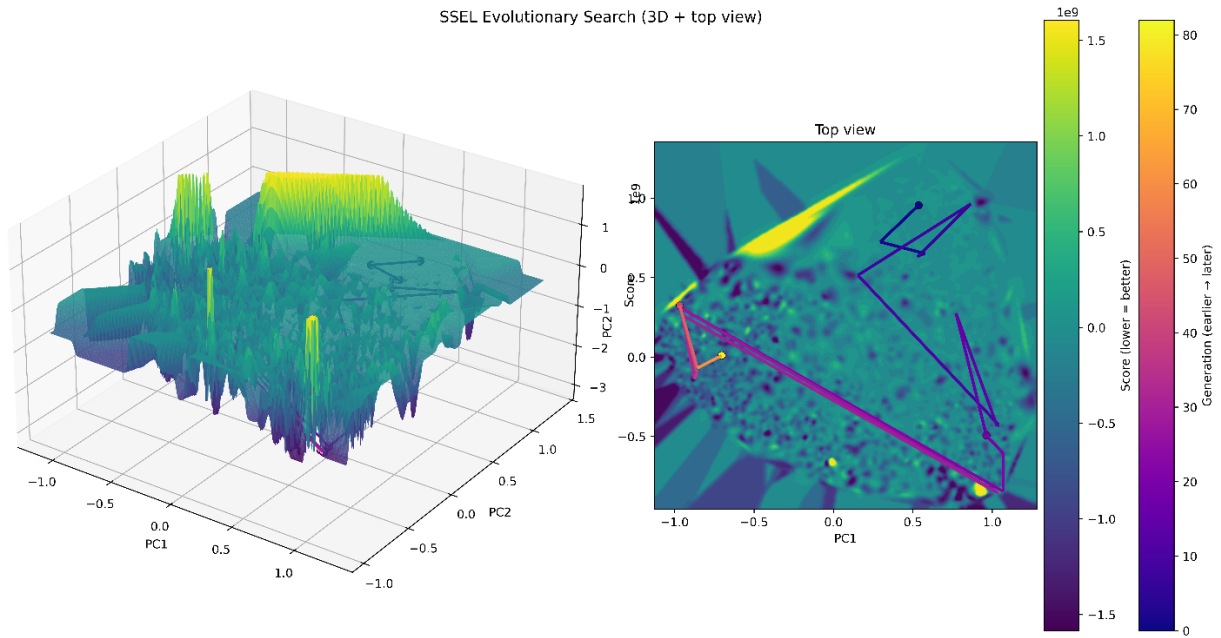

Figure 12. Joint 3D and top-view plots of the evolutionary search landscape for Subject B.

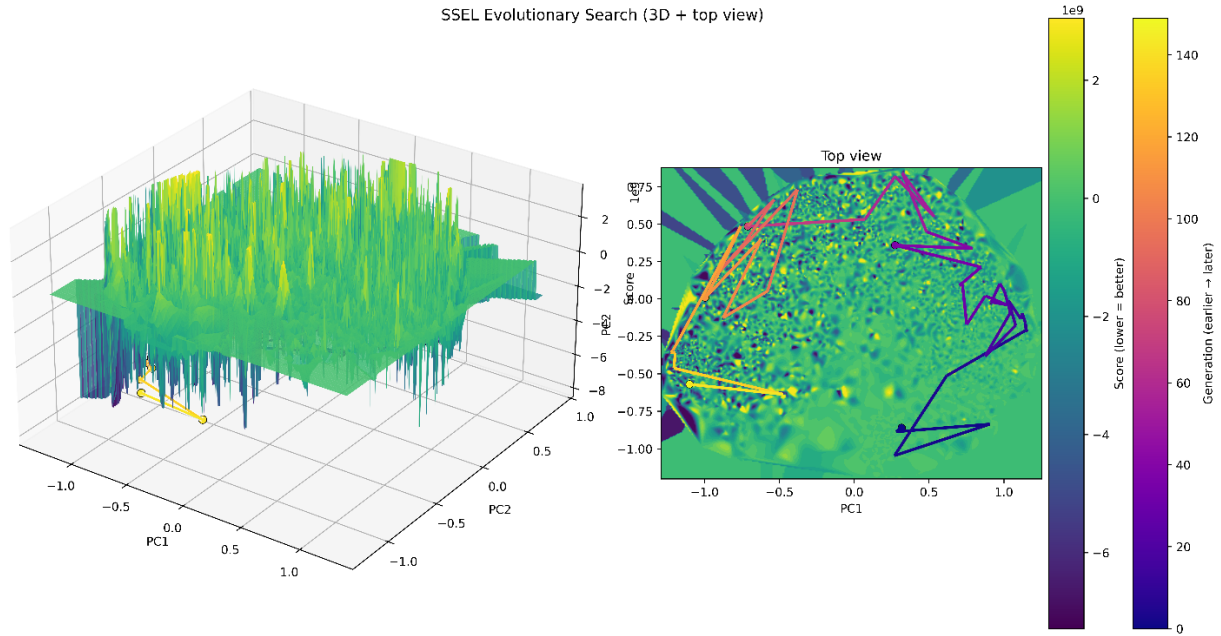

Figure 13. Joint 3D and top-view plots of the evolutionary search landscape for Subject B with a crossover rate of 0.2.

Furthermore, we systematically varied the mutation standard deviation (µ) and crossover rate (χ) to quantify their influence on convergence dynamics. Best-of-generation score curves were plotted across generations for each parameter configuration. For mutation, as shown in Figure 14, larger µ values promoted exploration and faster convergence, but they produced noisier trajectories and led to premature convergence. In contrast, smaller values (e.g., µ = 0.05) facilitated steady improvement and ultimately achieved the best overall performance. For crossover, moderate-to-high rates (χ = 0.60, χ = 0.80) generally produced faster convergence compared to lower crossover (χ = 0.40), which stagnated earlier. These patterns are consistent with the theoretical trade-off between exploration and exploitation: mutation injects diversity, while crossover accelerates the exploitation of promising solutions. Overall, these results demonstrate that the evolutionary search in our self-supervised framework is tunable, with optimal configurations balancing low mutation and high crossover to achieve both robustness and efficiency. And this sensitivity to evolutionary parameters highlights that discovering neurodynamic structure is not a deterministic optimization problem, but an exploration of a rugged landscape shaped by competing cognitive constraints such as stability, distinctness, and reproducibility.

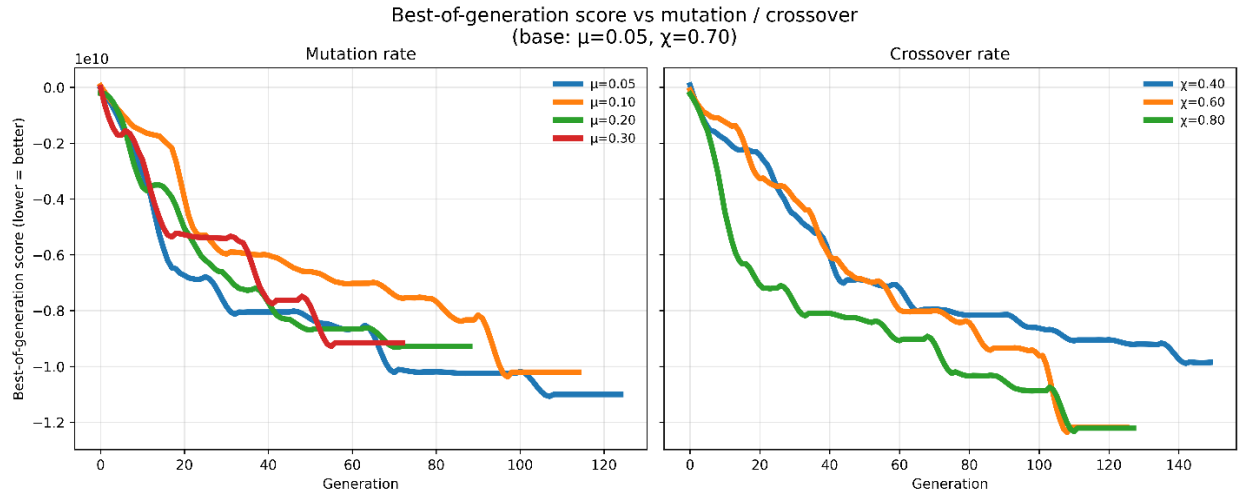

Figure 14. Comparison of best-of-generation score curves under varying mutation (left) and crossover (right) rates. Lower scores indicate better performance. Mutation governs exploration, whereas crossover governs exploitation speed.

### 4.4 Stability Analysis of Top Candidates

To evaluate the reliability of the segmentation, we quantified boundary stability using the normalized mean absolute deviation (MAD/T), which measures the variability of boundary placement across the top candidate solutions, scaled by trial length. This metric directly reflects how reproducibly the algorithm locates stage transitions under stochastic evolutionary search.

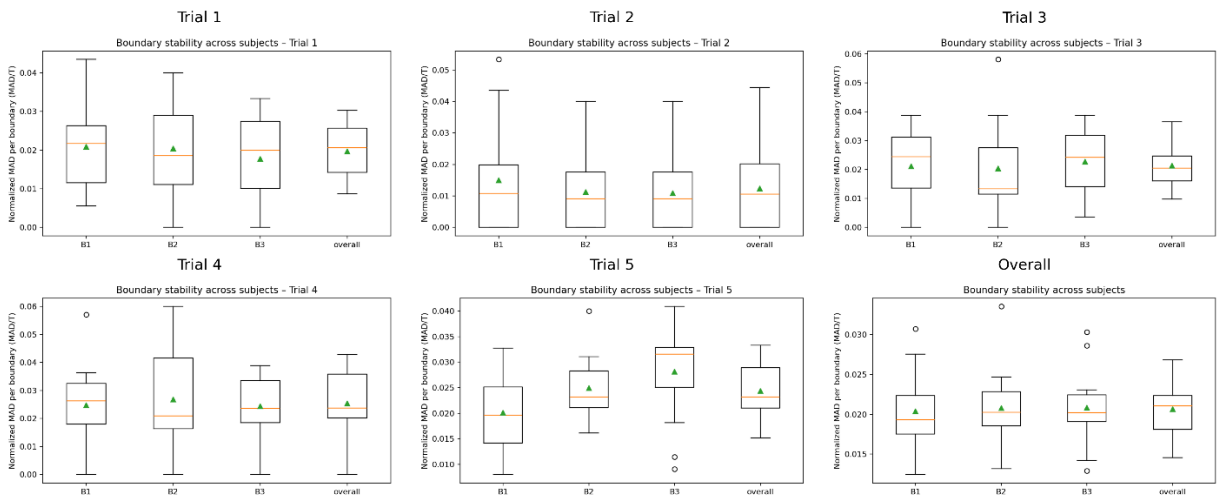

Figure 15. Boundary stability across subjects for five trials. Each subplot shows the distribution of normalized mean absolute deviation (MAD/T) of boundary positions for the three identified boundaries (B1, B2, B3) and the overall average. Lower MAD/T values indicate more consistent boundary placement across the top candidates.

A Friedman test across boundaries (B1, B2, B3, pooled across all trials) revealed no significant differences in stability ($\chi^2 = 0.081$, p = 0.96), indicating that the three boundaries were comparably stable and the algorithm was not biased toward one transition being noisier than others. A Kruskal–Wallis test across trials indicated only a trend toward differences in overall boundary stability ($\chi^2 = 9.29$, p = 0.054). Post-hoc Dunn's tests with Holm correction did not identify any significant pairwise differences. These results suggest that segmentation stability is consistent across both boundaries and trials, highlighting the generalizability of our method across diverse task contexts. In addition, the comparable stability across boundaries suggests that no single cognitive transition disproportionately dominates variability, supporting the interpretation that each stage transition represents a robust and recurrent cognitive event rather than a noisy artifact of the segmentation process.

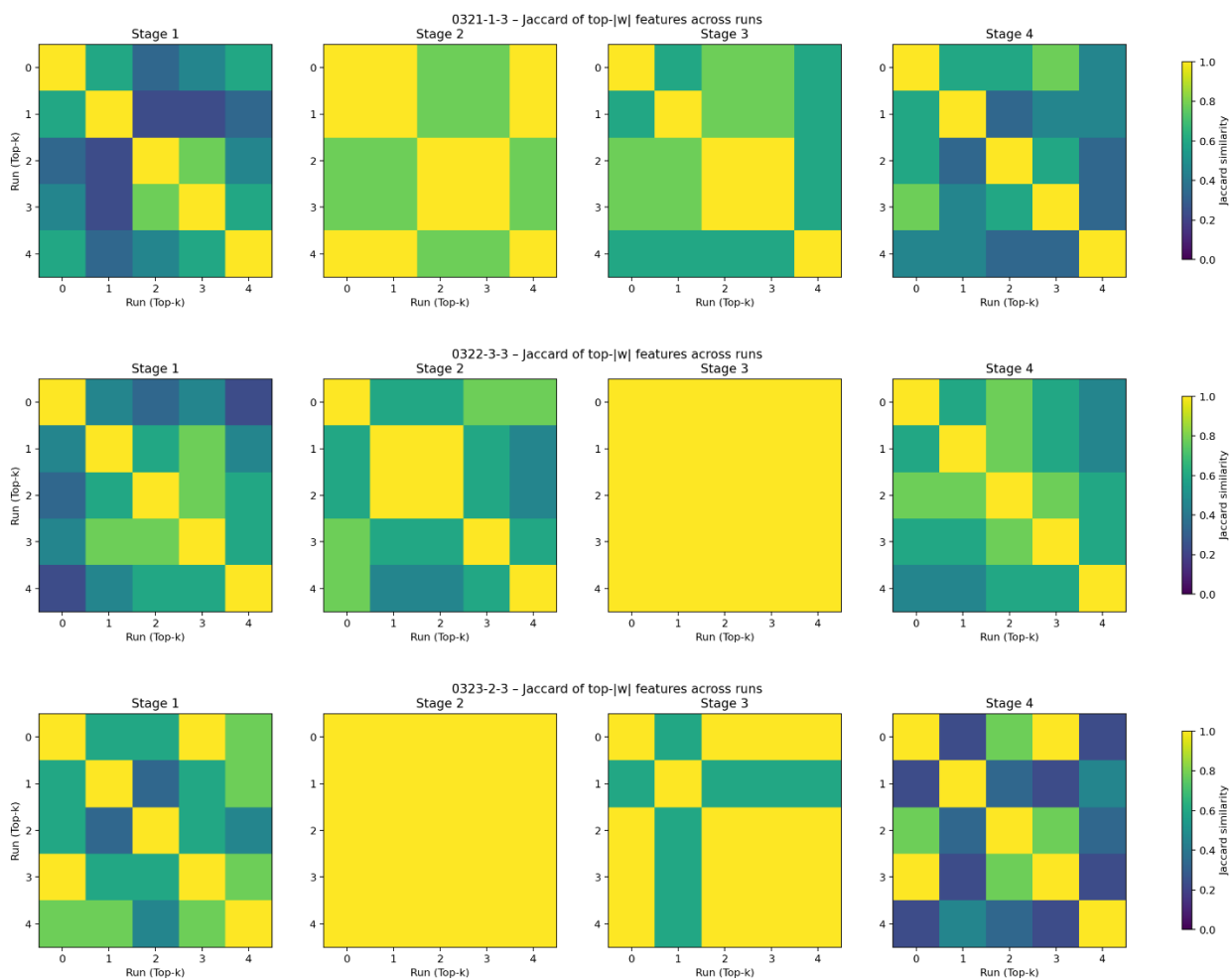

Figure 16. Heatmaps of Jaccard similarity for the top-8 features across top 5 candidates in each stage, shown for representative subjects. Bright yellow cells indicate high overlap (robust feature

convergence), whereas darker colors reflect variability. Heatmaps from all subjects can be found in Appendix.

To evaluate the stability of feature usage across candidates, we extracted the top-8 features (by absolute weight) from each candidate within a stage and computed the pairwise Jaccard similarity. The resulting heatmaps (Figure 16) quantify how consistently the top candidates selected the same features across runs. High Jaccard similarity across candidates means that the algorithm consistently converges on the same subset of EEG channels and frequency bands. This implies the representational structure extracted by SSEL is not an artifact of stochastic search, but reflects reproducible neural dynamics that generalize across runs.

Stages 2 and 3 showed high similarity, even approaching 1.0, indicating strong convergence toward a common representational basis. This suggests that evidence evaluation and motor readiness are associated with more canonical neural signatures, making them easier to capture robustly. This also provides confidence that these stages represent meaningful cognitive states rather than noise-driven partitions. The identified stable feature-channel pairs can inform future data collection by highlighting the most informative sensors and frequency bands. For instance, if FC5 BetaL consistently appears, future studies may prioritize denser sampling in central/motor regions. In practice, stable channel sets also mean that classifiers or control policies built on top of these features are less likely to overfit to specific runs, improving generalization to unseen data and new subjects.

By contrast, Stages 1 and 4 exhibited lower overlap, reflecting greater variability in how perceptual onset and execution-related dynamics were represented. This also suggests greater inter-individual flexibility in how brains implement these transitions, whereas the high convergence of features in Stages 2 and 3 indicates shared mechanisms.

These heatmaps can be seen as subject-level "fingerprints", and reflect genuine inter-individual differences in how neural features are recruited. For EEG-based applications, instead of assuming uniform transferability across all subjects, the heatmaps can provide a principled way to quantify "closeness" to guide transfer learning. If we want to transfer a model trained on one subject to another, we would ideally choose pairs with similar Jaccard similarity structures. In practice, we can cluster subjects by Jaccard similarity patterns to identify which individuals are likely to benefit from shared models and which need more personalized calibration.

Overall, these stability analyses demonstrate that SSEL recovers neurodynamic structures that are simultaneously robust across runs, interpretable at the cognitive level, and sensitive to individual differences, reinforcing its suitability for modeling both shared and personalized decision dynamics.

# 5 Discussion and Conclusion

## 5.1 Rationale for Evolutionary Search in Self-Supervised Neurodynamic Segmentation

Although SSL provides the foundation for mining temporal neurodynamic structure, the choice of evolutionary search over gradient-based optimization is central to the problem formulation addressed in this study. The primary unknowns in our framework are temporal boundaries $\tau^j = (t_1^j, t_2^j, t_3^j)$, which are discrete indices defining stage transitions in continuous EEG. These boundaries are inherently non-differentiable, and although continuous relaxations are possible, they tend to be unstable and sensitive to noise. Evolutionary search, however, naturally supports joint optimization over discrete and continuous variables without requiring differentiable approximations.

More fundamentally, our objective integrates multiple criteria (including within-stage predictability, boundary contrast, cross-trial alignment, and sparsity), many of which involve sub-model fitting, ranking operations, or hard selection mechanisms. These components yield a non-smooth and highly irregular optimization landscape that violates the assumption underlying gradient descent. Evolutionary search requires only the ability to evaluating candidate solutions, making it robust to non-differentiability and heterogeneous objective components.

Temporal segmentation of EEG further highly non-convex search space, where small shifts in boundary placement can lead to large changes in model fit or contrast measures. Gradient descent would easily get stuck in poor local optima. Population-based evolutionary search however maintain multiple candidate solutions in parallel, preserving diversity and enabling exploration across multiple basins of attraction. This global search capability is particularly important given the short trial durations, high noise levels, and inter-trial variability characteristic of EEG data.

Evolutionary search also enables interpretability via sparsity. The sparse weights $w_k$ are enforced by keeping only the top few features. This hard thresholding is non-differentiable. Gradient-based methods would need soft approximations (like Lasso), which however would blur interpretability. Evolutionary perturbation can directly turn features on/off, thus being able to yield clear electrode × band sets as transparent and analyzable neurodynamic signatures.

Finally, evolutionary search provides a flexible mechanism for incorporating behavioral priors (e.g. minimum segment length, trial alignment constraints) directly into the optimization process. These priors can be encoded as rules or constraints depending on the context of the real-world time-series. However, this is hard in gradient-based methods, unless these rules can be framed as differentiable penalties, which is not always possible.

## 5.2 Neurodynamic Interpretation and Cognitive Mechanisms

The stage structures and feature manifolds discovered by the proposed framework can be interpreted through the lens of perceptual decision-making as a neurodynamic trajectory. Across subjects and trials, the learned segmentations consistently revealed a progression that aligns with well-established phases of decision formation: sensory appraisal, evidence evaluation, motor readiness, and action execution or commitment. Importantly, these stages were not imposed a priori, but emerged endogenously from the temporal structure of EEG activity.

From a cognitive neurodynamics perspective, this finding supports the view that decision making unfolds as a movement through latent neural state space, where distinct processing stages correspond to metastable regions separated by relatively sharp transitions. Early stages are characterized by frontal–parietal engagement and broadband oscillatory activity associated with sensory integration and contextual appraisal, whereas intermediate stages exhibit signatures consistent with sustained evidence accumulation and top–down control. The emergence of a distinct motor readiness stage, marked by frontal theta enhancement and beta suppression, highlights a transitional regime in which action is prepared but withheld, a phenomenon widely reported in both human and animal studies of decision commitment.

Beyond the shared stage sequence, the learned identity manifolds revealed systematic individual differences in how these stages are instantiated. While the order of stages remained largely invariant across subjects, the duration, boundary sharpness, and feature composition of each stage varied. These differences can be interpreted as markers of cognitive style or decision-making phenotype, reflecting how individuals trade off speed, caution, and evidence requirements under uncertainty. For example, prolonged evaluative stages may correspond to more conservative decision strategies, whereas compressed intermediate stages may reflect rapid commitment once perceptual clarity is achieved.

At a higher theoretical level, the discovered trajectories resonate with dynamical systems accounts of cognition, in which neural activity evolves toward attractor-like states corresponding to decisions or actions. In this framing, cognitive processing is not defined by discrete symbolic operations, but by the system's trajectory through a structured landscape shaped by task demands and individual predispositions. The SSEL framework provides a computational instantiation of this idea by explicitly discovering the boundaries and manifolds that organize such trajectories from continuous neural data.

This interpretation reframes EEG not as a collection of features to be decoded, but as a temporal process that carries meaning in its evolution. By modeling progression itself as the primary object of learning, the proposed approach aligns closely with cognitive neurodynamic theories that emphasize time, structure, and internal state evolution as fundamental to cognition.

### 5.3 Connection to Cognitive Resilience, Redundancy and Quantum-Inspired Perspectives

Building on this neurodynamic interpretation, the variability observed across subjects can be understood not as noise, but as an expression of cognitive resilience and redundancy. As shown in our results, different subjects recruit partially different neurophysiological features even though they performed the same cognitive task and followed a similar decision-making progressive pattern. We acknowledge this as echoing the resilience and redundancy principles in the human mind, and, most importantly, design conceptually parallel methodology to accommodate this variability.

In the human brain, multiple redundant neural circuits often encode similar cognitive functions (e.g., motor planning distributed across frontal and parietal regions (Svoboda & Li, 2018)). This redundancy ensures resilience because if one pathway is disrupted (noise, fatigue, lesion), others can compensate (Stern et al., 2019). Cognitive resilience thus emerges from maintaining multiple internal hypotheses in parallel. Evolutionary search operates under an analogous principle. A population of candidate segmentations represents parallel hypotheses about temporal structure and feature relevance. Population diversity prevents premature convergence to suboptimal solutions (Sudholt, 2019), elitism preserves high-performing candidates, and mutation sustains exploration, which mirror how neural redundancy and variability support robust cognitive function.

This parallel also resonates with ideas from quantum-inspired models of parallelism, where multiple possibilities coexist until selection collapses the system toward a dominant state (Brody, 2023). Similarly, evolutionary populations maintain a distribution over candidate solutions, with selection acting to concentrate probability mass on higher-fitness structures while stochastic variation prevents overcommitment to a single trajectory. In both cases, redundancy and diversity serve as safeguards against noise and uncertainty.

Importantly, there principles are not merely pragmatic design choices for handling stochastic, non-stationary EEG signals. They reflect deeper mechanisms of cognitive resilience and uncertainty management. Selection pressure in evolutionary search functions as a form of hypothesis evaluation, analogous to the brain's error monitoring systems (anterior cingulate, prefrontal cortex) that assess whether ongoing cognitive states align with expected dynamics. In this sense, SSEL exhibits a rudimentary form of the metacognitive intelligence, i.e., maintaining, evaluating, and refining internal models of temporal structure to support robust and adaptive inference.

### 5.4 From Statistical Inference to Temporal Self-Reflection

The distinction between learning and statistical inference has long occupied a central place in both machine intelligence and neuroscience (Bzdok et al., 2018). Statistical inference detects whether a signal has shifted under predefined criteria, but remains hypothesis-driven, externally specified, and non-adaptive. Learning-based methods (including supervised, unsupervised, and reinforcement learning), in contrast, construct internal representations that improve with experience and generalize beyond fixed thresholds.

Most existing EEG segmentation methods fall into inference- or supervision-driven paradigms. Statistical methods detect change points on bandpower or functional connectivity features to delineate cognitive stages (Brodsky et al., 1999; Koerner et al., 2017), capturing distributional shifts but lacking an internal model of temporal progression. Supervised deep networks classify pre-labeled intervals and achieve strong discrimination (Lawhern et al., 2016; Saeidi et al., 2021), yet depend on costly annotations and often generalize poorly across subjects. Recent self-supervised approaches, primarily through contrastive objectives and applied for sleep staging (Lee et al., 2024; Yang et al., 2023; Zhao et al., 2024a), reduces labeling requirements, but largely follow a pretrain–finetune pipeline, where segmentation remains window-based and final stage definitions rely on human expert scoring. In these frameworks, self-supervision improves representation quality but does not endow the system with awareness of temporal progression.

In contrast to inference-based segmentation, the proposed framework enables a form of temporal self-reflection, in which the system models not only what state it is in, but how it arrived there through an evolving cognitive trajectory. Our framework departs from the previous paradigms by operating entirely without external labels and treating temporal boundary discovery as the primary objective. Rather than testing for change or classifying predefined states, the system learns stage transitions directly from intrinsic temporal structure, forming an internal representation of progression. This shift aligns more closely with theoretical neuroscience, where cognition emerges from neural dynamics (Durstewitz, 2003; Goel & Buonomano, 2014) rather than isolated inferential tests, and temporal sequence retention is a fundamental emergent computational ability (Hopfield, 1982).

The developed methodology also echoes the trend that evolve from external to internal reinforcement learning. Traditional reinforcement learning depends on explicit external rewards, but in many real-world scenarios, such as EEG-driven human–robot collaboration, such rewards are scarce or even unattainable. To address these limitations, researchers have developed intrinsic learning signals, such as curiosity (Pathak et al., 2017), prediction error (Burda et al., 2018), or entropy maximization (Liu & Abbeel, 2021). Agents trained to maximize entropy have learned locomotion skills (walking, jumping) without external guidance (Eysenbach et al., 2018), and intrinsic error-based rewards have enabled agents to surpass human performance in Atari games without demonstrations (Burda et al., 2018). These approaches align with principles of exploration and uncertainty reduction, echoing the "free-energy principle" in cognitive neuroscience (Friston, 2010). Yet they remain focused on exploring action spaces and optimizing tasks. Our method cultivates awareness of internal temporal evolution. The agent does not learn what to do next, but where it is in its own cognitive process.

### 5.5 Implications for Resilient and Progression-Aware Intelligent Systems

Metacognitive awareness of temporal progression opens new frontiers for artificial intelligence in smart civil and infrastructure systems. Most current applications rely on prediction-focused machine learning to forecast events or continuous variables. For example, in disaster, climate, and flooding management, supervised spatiotemporal modes are used to predict rainfall to flood extent (Xiao et al., 2025); in traffic monitoring, flow, congestion, and demand are forecasted (Afandizadeh et al., 2024). In complex, interacting real-world environments, however, directly mapping high-dimensional inputs to predictions is often brittle (D'Amour et al., 2022; Geirhos et al., 2020). Although attention-mechanisms partially address this limitation (Vaswani et al., 2017), their effectiveness depends on being context-aware, specifically, knowing which features are relevant at different stages; therefore, explicit stage-progression awareness is critical.

What many resilience and post-event analysis frameworks ultimately require is not sharper prediction, but self-reflection, i.e., the ability to recognize internal transitions toward vulnerability or failure. Our framework formalizes this capability by enabling models to infer temporal stage boundaries and dominant features directly from ten times higher-dimensional time-series data, without external supervision. Rather than labeling conditions or predicting future values, the system learns to identify tipping points and determine which features are most influential at each stage. This capability is particularly valuable in resilience-oriented settings, where preparedness is inherently staged (e.g., watch → warning → imminent failure), and early recognition of internal transitions is often more actionable than precise timing forecasts.

More broadly, today's smart systems implement the perception–action progression in a fragmented manner. For sensing (perception), smart systems gather data, such as IoT sensors, traffic cameras, flood gauges, structural health monitors. For processing (representation), they run predictive models (machine learning for traffic, hydrological models for flooding, risk indices for resilience). For actuation (action), responses like adaptive traffic lights, automated floodgates, smart HVAC adjustments, or transit re-routing are available. However, these stages are often decoupled and governed by low-level linear statistical inference, for example, threshold-based triggers (if water level > X, open gate; if congestion index > Y, reroute). The key advancement of our self-supervised, progression-aware framework is its ability to understand readiness in the perception-action loop from their own unfolding internal dynamics. It is not to replace the existing machine learning pipelines, but add an important layer of self-reflection for dynamically weighing based on internal transitions.

Another direct application of progression-aware learning is in EEG-based human–robot collaboration. Existing EEG-driven robotic systems typically decode neural signals into discrete commands such as pick or place, resume or stop (Zhou et al., 2025a). Due to the low signal-to-noise ratio of human EEG signals, additional mechanisms are often embedded to avoid false alarms; however, this can reduce sensitivity to genuine intent expression. Human decision-making, unfolds progressively: before action initiation, individuals often enter a motor readiness state in which action is prepared but withheld. Being able to track the progression of decision dynamics is particularly critical in complex or ambiguous situations. As demonstrated in this study, this readiness state is neurophysiologically distinct and temporally identifiable. A practical benefit can be illustrated in an EEG-driven robotic wheelchair at an intersection: recognizing that a user is prepared to move but waiting for a pedestrian to pass would allow the system to initiate motion immediately upon clearance, without requiring an explicit "resume" command. Such fine-grained intent awareness can significantly improve fluidity, safety, and user experience in BCI applications. This notion of progression awareness and readiness state also extends to autonomous driving, where current end-to-end pipelines often map sensory inputs directly to actuation controls such as steering angle or speed (Navarro et al., 2021). The absence of explicit intermediate readiness or progression states can render such systems fragile under unexpected scenarios. Incorporating progression-aware representations inspired by human motor planning may improve robustness and adaptability.

## 5.6 Generality of the Framework

Although the proposed framework is evaluated using a road-crossing task involving safety-critical decision-making, its formulation is not task-specific. SSEL does not rely on externally defined labels, task-specific annotations, or domain-dependent heuristics. Instead, it assumes only that cognition unfolds through temporally organized, stage-like neurodynamic regimes, an assumption shared by a wide range of cognitive processes. From a cognitive perspective, many tasks in experimental psychology and neuroscience exhibit similar progression structures. For example, perceptual decision-making tasks involving motion discrimination or signal detection typically involve stages of sensory evidence accumulation, uncertainty resolution, and commitment. Working memory tasks exhibit transitions between encoding, maintenance, and retrieval, each associated with distinct oscillatory and network-level signatures. Metacognitive confidence judgments involve additional stages in which internal representations are evaluated and revised. In principle, the same self-supervised evolutionary framework could be applied to continuous EEG collected during these tasks to discover task-specific yet structurally analogous neurodynamic progressions, without requiring predefined stage boundaries or trial annotations.

Beyond laboratory paradigms, the identity manifold perspective introduced in this work generalizes to settings where stable, person-specific neurodynamic signatures are critical. In brain–computer interfaces, such manifolds could support long-term personalization and adaptation by distinguishing enduring cognitive traits from transient states. In clinical monitoring, progression-aware segmentation may enable early detection of deviations from a patient's typical neurodynamic trajectory, providing individualized baselines for assessing cognitive fatigue, decline, or recovery. Importantly, these applications benefit from the same intrinsic properties demonstrated here: robustness to noise, resistance to spoofing, and interpretability of stage-specific neural signatures.

**Acknowledgments**

The work presented in this paper was supported financially by the United States National Science Foundation (NSF) via Award# SCC-IRG 2124857. The support of the NSF is gratefully acknowledged. Any opinions and findings in this paper are those of the authors and do not necessarily represent those of the NSF.

**Statements and Declarations**

## Competing Interests

The authors declare that they have no competing interests.

## References

Adams, R. P., & MacKay, D. J. (2007). Bayesian online changepoint detection. *arXiv preprint arXiv:0710.3742*.

Adobe. (2023). Creating a Motion tween animation. Retrieved from https://helpx.adobe.com/animate/using/creating_a_motion_tween_animation.html

Afandizadeh, S., Abdolahi, S., & Mirzahossein, H. (2024). Deep Learning Algorithms for Traffic Forecasting: A Comprehensive Review and Comparison with Classical Ones. *Journal of Advanced Transportation, 2024*(1), 9981657.

Aissa, N. E. H. S. B., Korichi, A., Lakas, A., Kerrache, C. A., & Calafate, C. T. (2024). Assessing robustness to adversarial attacks in attention-based networks: Case of EEG-based motor imagery classification. *SLAS technology, 29*(4), 100142.

Alahaideb, L., Al-Nafjan, A., Aljumah, H., & Aldayel, M. (2025). Brain-Computer Interface for EEG-Based Authentication: Advancements and Practical Implications. *Sensors (Basel), 25*(16).

Ali, A. S., Mahmood, S. A., Saeed, M., Konin, A., Zia, M. Z., & Tran, Q.-H. (2025). Joint Self-Supervised Video Alignment and Action Segmentation. *arXiv preprint arXiv:2503.16832*.

Aminikhanghahi, S., & Cook, D. J. (2017). A Survey of Methods for Time Series Change Point Detection. *Knowl Inf Syst, 51*(2), 339-367.

Arlot, S., Celisse, A., & Harchaoui, Z. (2019). A kernel multiple change-point algorithm via model selection. *Journal of Machine Learning Research, 20*(162), 1-56.

Baraku, T., Stergiadis, C., Veloudis, S., & Klados, M. A. (2024). Personalized user authentication system using wireless EEG headset and machine learning. *Brain Organoid and Systems Neuroscience Journal, 2*, 17-22.

Bermant, P. C., Brickson, L., & Titus, A. J. (2022). Bioacoustic event detection with self-supervised contrastive learning. *bioRxiv*, 2022.2010. 2012.511740.

Bhati, S., Villalba, J., Żelasko, P., Moro-Velazquez, L., & Dehak, N. (2022). Unsupervised speech segmentation and variable rate representation learning using segmental contrastive predictive coding. *IEEE/ACM Transactions on Audio, Speech, and Language Processing, 30*, 2002-2014.

Borst, J. P., & Anderson, J. R. (2015). The discovery of processing stages: analyzing EEG data with hidden semi-Markov models. *Neuroimage, 108*, 60-73.

Brill, S., Kumar Debnath, A., Payre, W., Horan, B., & Birrell, S. (2024). Factors influencing the perception of safety for pedestrians and cyclists through interactions with automated vehicles in shared spaces. *Transportation Research Part F: Traffic Psychology and Behaviour, 107*, 181-195.

Brodsky, B. E., Darkhovsky, B. S., Kaplan, A. Y., & Shishkin, S. L. (1999). A nonparametric method for the segmentation of the EEG. *Comput Methods Programs Biomed, 60*(2), 93-106.

Brody, D. C. (2023). Quantum formalism for the dynamics of cognitive psychology. *Scientific Reports, 13*(1), 16104.

Burda, Y., Edwards, H., Storkey, A., & Klimov, O. (2018). Exploration by random network distillation. *arXiv preprint arXiv:1810.12894*.

Bzdok, D., Altman, N., & Krzywinski, M. (2018). Statistics versus machine learning. *Nature methods, 15*(4), 233-234.

Chan, H.-L., Kuo, P.-C., Cheng, C.-Y., & Chen, Y.-S. (2018). Challenges and Future Perspectives on Electroencephalogram-Based Biometrics in Person Recognition. *Frontiers in Neuroinformatics, Volume 12 - 2018*.

Chen, X., Jia, T., & Wu, D. (2025). Data alignment based adversarial defense benchmark for EEG-based BCIs. *Neural Netw., 188*(C), 25.

Chen, X., Meng, L., Xu, Y., & Wu, D. (2024). Adversarial artifact detection in EEG-based brain–computer interfaces. *Journal of Neural Engineering, 21*(5), 056043.

Chen, X., Sun, Y., Zhang, M., & Peng, D. (2020). Evolving deep convolutional variational autoencoders for image classification. *IEEE Transactions on Evolutionary Computation, 25*(5), 815-829.

Chien, H.-Y. S., Goh, H., Sandino, C. M., & Cheng, J. Y. (2022). Maeeg: Masked auto-encoder for eeg representation learning. *arXiv preprint arXiv:2211.02625*.

Cochran, W., Cooley, J., Favin, D., Helms, H., Kaenel, R., Lang, W., Jr, G. C., Nelson, D., Rader, C., & Welch, P. (1967). What is the fast Fourier transform? *Proceedings of the IEEE, 15*, 1664-1674.

D'Amour, A., Heller, K., Moldovan, D., Adlam, B., Alipanahi, B., Beutel, A., Chen, C., Deaton, J., Eisenstein, J., & Hoffman, M. D. (2022). Underspecification presents challenges for credibility in modern machine learning. *Journal of Machine Learning Research, 23*(226), 1-61.

Dai, R.-J., Hu, K., Yin, H., Lu, B.-L., & Zheng, W.-L. (2025). *Self-Supervised EEG Representation Learning Based on Temporal Prediction and Spatial Reconstruction for Emotion Recognition.* Paper presented at the Proceedings of the Annual Meeting of the Cognitive Science Society.

De Clercq, K., Dietrich, A., Núñez Velasco, J. P., De Winter, J., & Happee, R. (2019). External human-machine interfaces on automated vehicles: Effects on pedestrian crossing decisions. *Human Factors, 61*(8), 1353-1370.

Debie, E., Moustafa, N., & Whitty, M. T. (2020). *A privacy-preserving generative adversarial network method for securing EEG brain signals.* Paper presented at the 2020 International Joint Conference on Neural Networks (IJCNN).

Deldari, S., Xue, H., Saeed, A., He, J., Smith, D. V., & Salim, F. D. (2022). Beyond just vision: A review on self-supervised representation learning on multimodal and temporal data. *arXiv preprint arXiv:2206.02353*.

Dubey, A., Markowitz, D. A., & Pesaran, B. (2023). Top-down control of exogenous attentional selection is mediated by beta coherence in prefrontal cortex. *Neuron, 111*(20), 3321-3334.e3325.

Durstewitz, D. (2003). Self-organizing neural integrator predicts interval times through climbing activity. *Journal of Neuroscience, 23*(12), 5342-5353.

EMOTIV. (2023). EmotivPRO v3.0. Retrieved from https://emotiv.gitbook.io/emotivpro-v3/data-streams/frequency-bands

Eysenbach, B., Gupta, A., Ibarz, J., & Levine, S. (2018). Diversity is all you need: Learning skills without a reward function. *arXiv preprint arXiv:1802.06070*.

Foxe, J., & Snyder, A. (2011). The Role of Alpha-Band Brain Oscillations as a Sensory Suppression Mechanism during Selective Attention. *Frontiers in Psychology, 2*(154).
Friston, K. (2010). The free-energy principle: a unified brain theory? *Nature Reviews Neuroscience, 11*(2), 127-138.
Fu, T., Hu, W., Miranda-Moreno, L., & Saunier, N. (2019). Investigating secondary pedestrian-vehicle interactions at non-signalized intersections using vision-based trajectory data. *Transportation Research Part C: Emerging Technologies, 105*, 222-240.
Gee, A. H., Chang, J., Ghosh, J., & Paydarfar, D. (2018). Bayesian Online Changepoint Detection Of Physiological Transitions. *Annu Int Conf IEEE Eng Med Biol Soc, 2018*, 45-48.
Geirhos, R., Jacobsen, J.-H., Michaelis, C., Zemel, R., Brendel, W., Bethge, M., & Wichmann, F. A. (2020). Shortcut learning in deep neural networks. *Nature Machine Intelligence, 2*(11), 665-673.
Gerogiannis, A., & Bode, N. W. F. (2024). Analysis of long-term observational data on pedestrian road crossings at unmarked locations. *Safety Science, 172*, 106420.
Goel, A., & Buonomano, D. V. (2014). Timing as an intrinsic property of neural networks: evidence from in vivo and in vitro experiments. *Philosophical Transactions of the Royal Society B: Biological Sciences, 369*(1637), 20120460.
Gramfort, A., Luessi, M., Larson, E., Engemann, D. A., Strohmeier, D., Brodbeck, C., Goj, R., Jas, M., Brooks, T., Parkkonen, L., & Hämäläinen, M. (2013). MEG and EEG data analysis with MNE-Python. *Frontiers in Neuroscience, 7*.
Gretton, A., Borgwardt, K. M., Rasch, M. J., Schölkopf, B., & Smola, A. (2012). A kernel two-sample test. *The journal of machine learning research, 13*(1), 723-773.
Habashi, A. G., Azab, A. M., Eldawlatly, S., & Aly, G. M. (2023). Generative adversarial networks in EEG analysis: an overview. *J Neuroeng Rehabil, 20*(1), 40.
Harchaoui, Z., Moulines, E., & Bach, F. (2008). Kernel change-point analysis. *Advances in Neural Information Processing Systems, 21*.
Hopfield, J. J. (1982). Neural networks and physical systems with emergent collective computational abilities. *Proceedings of the National Academy of Sciences, 79*(8), 2554-2558.
Hu, K., Dai, R.-J., Chen, W.-T., Yin, H.-L., Lu, B.-L., & Zheng, W.-L. (2024). *Contrastive Self-supervised EEG Representation Learning for Emotion Classification.* Paper presented at the 2024 46th Annual International Conference of the IEEE Engineering in Medicine and Biology Society (EMBC).
Huang, Y., Chen, Y., Xu, S., Wu, D., & Wu, X. (2025). Self-Supervised Learning with Adaptive Frequency-Time Attention Transformer for Seizure Prediction and Classification. *Brain Sciences, 15*(4), 382. doi:10.3390/brainsci15040382
Jalaly Bidgoly, A., Jalaly Bidgoly, H., & Arezoumand, Z. (2020). A survey on methods and challenges in EEG based authentication. *Computers & Security, 93*, 101788.
Jorge, J., van der Zwaag, W., & Figueiredo, P. (2014). EEG–fMRI integration for the study of human brain function. *Neuroimage, 102*, 24-34.
Karakaş, S. (2020). A review of theta oscillation and its functional correlates. *International Journal of Psychophysiology*.
Khalil, A. E., Perez-Diaz, J. A., Cantoral-Ceballos, J. A., & Antelis, J. M. (2024). Unlocking Security for Comprehensive Electroencephalogram-Based User Authentication Systems. *Sensors, 24*(24), 7919. doi:10.3390/s24247919

Killick, R., Fearnhead, P., & Eckley, I. A. (2012). Optimal Detection of Changepoints With a Linear Computational Cost. *Journal of the American Statistical Association, 107*(500), 1590-1598.

Klem, G. H., Lüders, H. O., Jasper, H. H., & Elger, C. (1999). The ten-twenty electrode system of the International Federation. The International Federation of Clinical Neurophysiology. *Electroencephalogr Clin Neurophysiol Suppl, 52*, 3-6.

Koerner, F. S., Anderson, J. R., Fincham, J. M., & Kass, R. E. (2017). Change-point detection of cognitive states across multiple trials in functional neuroimaging. *Stat Med, 36*(4), 618-642.

Kostas, D., Aroca-Ouellette, S., & Rudzicz, F. (2021). BENDR: Using transformers and a contrastive self-supervised learning task to learn from massive amounts of EEG data. *Frontiers in Human Neuroscience, 15*, 653659.

Kreuk, F., Keshet, J., & Adi, Y. (2020). Self-supervised contrastive learning for unsupervised phoneme segmentation. *arXiv preprint arXiv:2007.13465*.

Lander, S., & Shang, Y. (2015). *EvoAE--a new evolutionary method for training autoencoders for deep learning networks.* Paper presented at the 2015 IEEE 39th annual computer software and applications conference.

Lawhern, V., Solon, A., Waytowich, N., Gordon, S., Hung, C., & Lance, B. (2016). EEGNet: A Compact Convolutional Network for EEG-based Brain-Computer Interfaces. *Journal of Neural Engineering, 15*.

Lee, S., Yu, Y., Back, S., Seo, H., & Lee, K. (2024). SleePyCo: Automatic sleep scoring with feature pyramid and contrastive learning. *Expert Systems With Applications, 240*, 122551.

Li, Y., Chen, J., Li, F., Fu, B., Wu, H., Ji, Y., Zhou, Y., Niu, Y., Shi, G., & Zheng, W. (2022). GMSS: Graph-based multi-task self-supervised learning for EEG emotion recognition. *Ieee Transactions on Affective Computing, 14*(3), 2512-2525.

Li, Y., Hao, C., Li, P., Xiong, J., & Chen, D. (2021). Generic neural architecture search via regression. *Advances in Neural Information Processing Systems, 34*, 20476-20490.

Liu, H., & Abbeel, P. (2021). *Aps: Active pretraining with successor features.* Paper presented at the International Conference on Machine Learning.

Liu, Y., Xue, W., Yang, L., & Li, M. (2025). Deep Learning-Based EEG Emotion Recognition: A Review. *Brain Sci, 16*(1).

Luo, T. Z., Kim, T. D., Gupta, D., Bondy, A. G., Kopec, C. D., Elliott, V. A., DePasquale, B., & Brody, C. D. (2025). Transitions in dynamical regime and neural mode during perceptual decisions. *Nature, 646*(8087), 1156-1166.

Mahini, R., Li, Y., Ding, W., Fu, R., Ristaniemi, T., Nandi, A. K., Chen, G., & Cong, F. (2020). Determination of the Time Window of Event-Related Potential Using Multiple-Set Consensus Clustering. *Front Neurosci, 14*, 521595.

Maidstone, R., Hocking, T., Rigaill, G., & Fearnhead, P. (2017). On optimal multiple changepoint algorithms for large data. *Stat Comput, 27*(2), 519-533.

Meng, L., Jiang, X., Chen, X., Liu, W., Luo, H., & Wu, D. (2024). Adversarial filtering based evasion and backdoor attacks to EEG-based brain-computer interfaces. *Information Fusion, 107*, 102316.

Meng, L., Jiang, X., Huang, J., Zeng, Z., Yu, S., Jung, T.-P., Lin, C.-T., Chavarriaga, R., & Wu, D. (2023). EEG-based brain–computer interfaces are vulnerable to backdoor attacks. *Ieee Transactions on Neural Systems and Rehabilitation Engineering, 31*, 2224-2234.

Mohsenvand, M. N., Izadi, M. R., & Maes, P. (2020). *Contrastive representation learning for electroencephalogram classification.* Paper presented at the Machine learning for health.

Navarro, P. J., Miller, L., Rosique, F., Fernández-Isla, C., & Gila-Navarro, A. (2021). End-to-End Deep Neural Network Architectures for Speed and Steering Wheel Angle Prediction in Autonomous Driving. *Electronics, 10*(11), 1266. doi:10.3390/electronics10111266

Ou, Y., Sun, S., Gan, H., Zhou, R., & Yang, Z. (2022). An improved self-supervised learning for EEG classification. *Math. Biosci. Eng, 19*(7), 6907-6922.

Parr, T., Friston, K., & Pezzulo, G. (2024). Generative models for sequential dynamics in active inference. *Cognitive Neurodynamics, 18*(6), 3259-3272.

Pathak, D., Agrawal, P., Efros, A. A., & Darrell, T. (2017). *Curiosity-driven exploration by self-supervised prediction.* Paper presented at the International conference on machine learning.

Peirce, J., Gray, J. R., Simpson, S., MacAskill, M., Höchenberger, R., Sogo, H., Kastman, E., & Lindeløv, J. K. (2019). PsychoPy2: Experiments in behavior made easy. *Behavior Research Methods, 51*(1), 195-203.

Peter, J., Ferraioli, F., Mathew, D., George, S., Chan, C., Alalade, T., Salcedo, S. A., Saed, S., Tatti, E., Quartarone, A., & Ghilardi, M. F. (2022). Movement-related beta ERD and ERS abnormalities in neuropsychiatric disorders. *Frontiers in Neuroscience, Volume 16 - 2022*.

Rafiei, M. H., Gauthier, L. V., Adeli, H., & Takabi, D. (2022). Self-supervised learning for electroencephalography. *IEEE Transactions on Neural Networks and Learning Systems, 35*(2), 1457-1471.

Rehman, T. U., Alruwaili, M., Siddiqi, M. H., Alhwaiti, Y., Anwar, S., Halim, Z., & Alam, M. (2025). Advancing EEG-based biometric identification through multi-modal data fusion and deep learning techniques. *Complex & Intelligent Systems, 11*(9), 398.

Rivas-Carrillo, S. D., Akkuratov, E. E., Valdez Ruvalcaba, H., Vargas-Sanchez, A., Komorowski, J., San-Juan, D., & Grabherr, M. G. (2023). MindReader: Unsupervised Classification of Electroencephalographic Data. *Sensors (Basel), 23*(6).

Saeidi, M., Karwowski, W., Farahani, F. V., Fiok, K., Taiar, R., Hancock, P. A., & Al-Juaid, A. (2021). Neural Decoding of EEG Signals with Machine Learning: A Systematic Review. *Brain Sci, 11*(11).

Saibene, A., Ghaemi, H., & Dagdevir, E. (2024). Deep learning in motor imagery EEG signal decoding: A Systematic Review. *Neurocomputing, 610*, 128577.

Sgouralis, I., & Pressé, S. (2017). An Introduction to Infinite HMMs for Single-Molecule Data Analysis. *Biophysical Journal, 112*(10), 2021-2029.

Shen, X., Tao, L., Chen, X., Song, S., Liu, Q., & Zhang, D. (2024). Contrastive learning of shared spatiotemporal EEG representations across individuals for naturalistic neuroscience. *Neuroimage, 301*, 120890.

Siddiqi, S. M., & Moore, A. W. (2005). *Fast inference and learning in large-state-space HMMs.* Paper presented at the Proceedings of the 22nd international conference on Machine learning.

Sinn, M., Ghodsi, A., & Keller, K. (2012a). Detecting change-points in time series by maximum mean discrepancy of ordinal pattern distributions. *arXiv preprint arXiv:1210.4903*.

Sinn, M., Ghodsi, A., & Keller, K. (2012b). *Detecting change-points in time series by maximum mean discrepancy of ordinal pattern distributions*. Paper presented at the Proceedings of the Twenty-Eighth Conference on Uncertainty in Artificial Intelligence, Catalina Island, CA.
Song, A. H., Chlon, L., Soulat, H., Tauber, J., Subramanian, S., Ba, D., & Prerau, M. J. (2019). Multitaper Infinite Hidden Markov Model for EEG. *Annu Int Conf IEEE Eng Med Biol Soc, 2019*, 5803-5807.
Stergiadis, C., Kostaridou, V.-D., Veloudis, S., Kazis, D., & Klados, M. A. (2022). A personalized user authentication system based on EEG signals. *Sensors, 22*(18), 6929.
Stern, Y., Barnes, C. A., Grady, C., Jones, R. N., & Raz, N. (2019). Brain reserve, cognitive reserve, compensation, and maintenance: operationalization, validity, and mechanisms of cognitive resilience. *Neurobiol Aging, 83*, 124-129.
Sudholt, D. (2019). The benefits of population diversity in evolutionary algorithms: a survey of rigorous runtime analyses. *Theory of evolutionary computation: Recent developments in discrete optimization*, 359-404.
Sun, R., Zhuang, X., Wu, C., Zhao, G., & Zhang, K. (2015). The estimation of vehicle speed and stopping distance by pedestrians crossing streets in a naturalistic traffic environment. *Transportation Research Part F: Traffic Psychology and Behaviour, 30*, 97-106.
Sun, Y., Yen, G. G., & Yi, Z. (2018). Evolving unsupervised deep neural networks for learning meaningful representations. *IEEE Transactions on Evolutionary Computation, 23*(1), 89-103.
Svoboda, K., & Li, N. (2018). Neural mechanisms of movement planning: motor cortex and beyond. *Current Opinion in Neurobiology, 49*, 33-41.
Tian, K., Markkula, G., Wei, C., Lee, Y. M., Madigan, R., Hirose, T., Merat, N., & Romano, R. (2024). Deconstructing Pedestrian Crossing Decisions in Interactions With Continuous Traffic: An Anthropomorphic Model. *Trans. Intell. Transport. Sys., 25*(3), 2466–2478.
Tognoli, E., & Kelso, J. A. (2014). The metastable brain. *Neuron, 81*(1), 35-48.
Truong, C., Oudre, L., & Vayatis, N. (2018). ruptures: change point detection in Python. *arXiv preprint arXiv:1801.00826*.
Truong, C., Oudre, L., & Vayatis, N. (2020). Selective review of offline change point detection methods. *Signal Processing, 167*, 107299.
Truong, Q.-T., Nguyen, D. T., Hua, B.-S., & Yeung, S.-K. (2024). Self-supervised video object segmentation with distillation learning of deformable attention. *arXiv preprint arXiv:2401.13937*.
Tucci, C., Biasi, L. D., Marco, F. D., Citarella, A. A., Greca, A. D., Amaro, I., & Tortora, G. (2025). *Toward EEG Based Biometric Authentication in VirtualReality: a Pilot Usability and Performance Study*. Paper presented at the Proceedings of the 16th Biannual Conference of the Italian SIGCHI Chapter. https://doi.org/10.1145/3750069.3750093
Vadher, H., Patel, P., Nair, A., Vyas, T., Desai, S., Gohil, L., Tanwar, S., Garg, D., & Singh, A. (2024). EEG-based biometric authentication system using convolutional neural network for military applications. *SECURITY AND PRIVACY, 7*(2), e345.
Vaswani, A., Shazeer, N., Parmar, N., Uszkoreit, J., Jones, L., Gomez, A. N., Kaiser, Ł., & Polosukhin, I. (2017). Attention is all you need. *Advances in Neural Information Processing Systems, 30*.
Vinhas, A., Correia, J., & Machado, P. (2025). Evolutionary Machine Learning Meets Self-Supervised Learning: A Comprehensive Survey. *arXiv preprint arXiv:2504.07213*.

Wagner, J., Wessel, J., Ghahremani, A., & Aron, A. (2017). Establishing a Right Frontal Beta Signature for Stopping Action in Scalp Electroencephalography: Implications for Testing Inhibitory Control in Other Task Contexts. *Journal of Cognitive Neuroscience, 30*, 1-12.

Wang, E. T., Vannucci, M., Haneef, Z., Moss, R., Rao, V. R., & Chiang, S. (2022a). A Bayesian switching linear dynamical system for estimating seizure chronotypes. *Proceedings of the National Academy of Sciences, 119*(46), e2200822119.

Wang, G., Liu, W., He, Y., Xu, C., Ma, L., & Li, H. (2024). Eegpt: Pretrained transformer for universal and reliable representation of eeg signals. *Advances in Neural Information Processing Systems, 37*, 39249-39280.

Wang, M., Wang, S., & Hu, J. (2022b). Cancellable template design for privacy-preserving EEG biometric authentication systems. *IEEE Transactions on Information Forensics and Security, 17*, 3350-3364.

Wang, X. J. (2012). Neural dynamics and circuit mechanisms of decision-making. *Curr Opin Neurobiol, 22*(6), 1039-1046.

Wang, Y., Srinivasan, A. R., Jokinen, J. P. P., Oulasvirta, A., & Markkula, G. (2025). Pedestrian crossing decisions can be explained by bounded optimal decision-making under noisy visual perception. *Transportation Research Part C: Emerging Technologies, 171*, 104963.

Wang, Z., Chen, H., Li, X., Liu, C., Xiong, Y., Tighe, J., & Fowlkes, C. (2022c). *Sscap: Self-supervised co-occurrence action parsing for unsupervised temporal action segmentation.* Paper presented at the Proceedings of the IEEE/CVF Winter Conference on Applications of Computer Vision.

Weng, W., Gu, Y., Guo, S., Ma, Y., Yang, Z., Liu, Y., & Chen, Y. (2025). Self-supervised learning for electroencephalogram: A systematic survey. *ACM Computing Surveys, 57*(12), 1-38.

Williams, N. J., Daly, I., & Nasuto, S. J. (2018). Markov Model-Based Method to Analyse Time-Varying Networks in EEG Task-Related Data. *Frontiers in Computational Neuroscience, Volume 12 - 2018*.

Wu, D., Xu, J., Fang, W., Zhang, Y., Yang, L., Xu, X., Luo, H., & Yu, X. (2023). Adversarial attacks and defenses in physiological computing: a systematic review. *Natl Sci Open, 2*(1).

Xia, K., Duch, W., Sun, Y., Xu, K., Fang, W., Luo, H., Zhang, Y., Sang, D., Xu, X., & Wang, F.-Y. (2022). Privacy-preserving brain–computer interfaces: A systematic review. *IEEE Transactions on Computational Social Systems, 10*(5), 2312-2324.

Xiao, J., Wang, Z., Liao, Y., Yi, Y., Zheng, L., Yang, B., Yu, H., Li, X., Hu, N., & Lai, C. (2025). A ConvLSTM-Based Model for Urban Flood Prediction Under Dynamic Rainfall Patterns and Exploration on Its Extrapolation Capability. *International Journal of Disaster Risk Science, 16*(6), 1057-1073.

Yang, B., Zhu, X., Liu, Y., & Liu, H. (2021a). A single-channel EEG based automatic sleep stage classification method leveraging deep one-dimensional convolutional neural network and hidden Markov model. *BIOMEDICAL SIGNAL PROCESSING AND CONTROL, 68*, 102581.

Yang, C., Lamdouar, H., Lu, E., Zisserman, A., & Xie, W. (2021b). *Self-supervised video object segmentation by motion grouping.* Paper presented at the Proceedings of the IEEE/CVF International Conference on Computer Vision.

Yang, C., Xiao, C., Westover, M. B., & Sun, J. (2023). Self-supervised electroencephalogram representation learning for automatic sleep staging: model development and evaluation study. *JMIR AI, 2*(1), e46769.

Yao, J., Hong, Y., Wang, C., Xiao, T., He, T., Locatello, F., Wipf, D. P., Fu, Y., & Zhang, Z. (2022). Self-supervised amodal video object segmentation. *Advances in Neural Information Processing Systems, 35*, 6278-6291.

Yıldız, İ., Garner, R., Lai, M., & Duncan, D. (2022). Unsupervised seizure identification on EEG. *Comput Methods Programs Biomed, 215*, 106604.

Yue, L., Tran, T. T. M., Yu, X., & Hoggenmueller, M. (2025). *Enhancing Autonomous Vehicle-Pedestrian Interaction in Shared Spaces: The Impact of Intended Path-Projection*. Paper presented at the Proceedings of the Extended Abstracts of the CHI Conference on Human Factors in Computing Systems. https://doi.org/10.1145/3706599.3720174

Zhang, J., Yang, L., Sajjadi Mohammadabadi, S. M., & Yan, F. (2025a). A survey on self-supervised learning: Recent advances and open problems. *Neurocomputing, 655*, 131409.

Zhang, M., Liu, D., Hu, S., Yan, X., Sun, Z., & Ye, Y. (2023). Self-supervised temporal autoencoder for egocentric action segmentation. *Engineering Applications of Artificial Intelligence, 126*, 107092.

Zhang, X., Wu, D., Ding, L., Luo, H., Lin, C.-T., Jung, T.-P., & Chavarriaga, R. (2020). Tiny noise, big mistakes: adversarial perturbations induce errors in brain–computer interface spellers. *National Science Review, 8*(4).

Zhang, Z., Elahi, M. F., Domeyer, J., & Tian, R. (2025b). Driver temporal segmentation of pedestrian crossing intentions during negotiations. *Transportation Research Part F: Traffic Psychology and Behaviour, 114*, 953-969.

Zhao, C., Wu, W., Zhang, H., Zhang, R., Zheng, X., & Kong, X. (2024a). Sleep Stage Classification Via Multi-View Based Self-Supervised Contrastive Learning of EEG. *IEEE J Biomed Health Inform, 28*(12), 7068-7077.

Zhao, T., Cui, Y., Ji, T., Luo, J., Li, W., Jiang, J., Gao, Z., Hu, W., Yan, Y., Jiang, Y., & Hong, B. (2024b). VAEEG: Variational auto-encoder for extracting EEG representation. *Neuroimage, 304*, 120946.

Zhou, X., & Liao, P.-C. (2022). A privacy-preserving data storage and service framework based on deep learning and blockchain for construction workers' wearable IoT sensors. *arXiv preprint arXiv:2211.10713*.

Zhou, X., Menassa, C. C., & Kamat, V. R. (2024). Decoding Brain Dynamics in Motor Planning Based on EEG Microstates for Predicting Pedestrian Road-Crossing in Vehicle-to-Everything Architectures. *arXiv preprint arXiv:2405.13955*.

Zhou, X., Menassa, C. C., & Kamat, V. R. (2025a). Feasibility of Embodied Dynamics Based Bayesian Learning for Continuous Pursuit Motion Control of Assistive Mobile Robots in the Built Environment. *arXiv preprint arXiv:2511.17401*.

Zhou, X., Menassa, C. C., & Kamat, V. R. (2025b). Siamese network with dual attention for EEG-driven social learning: Bridging the human-robot gap in long-tail autonomous driving. *Expert Systems With Applications, 291*, 128470.

Zou, H., Guo, Y., Wei, F., Guo, D., Li, Q., & Pirov, J. (2025). A pedestrian group crossing intention prediction model integrating spatiotemporal features. *Scientific Reports, 15*(1), 20675.

# Appendix

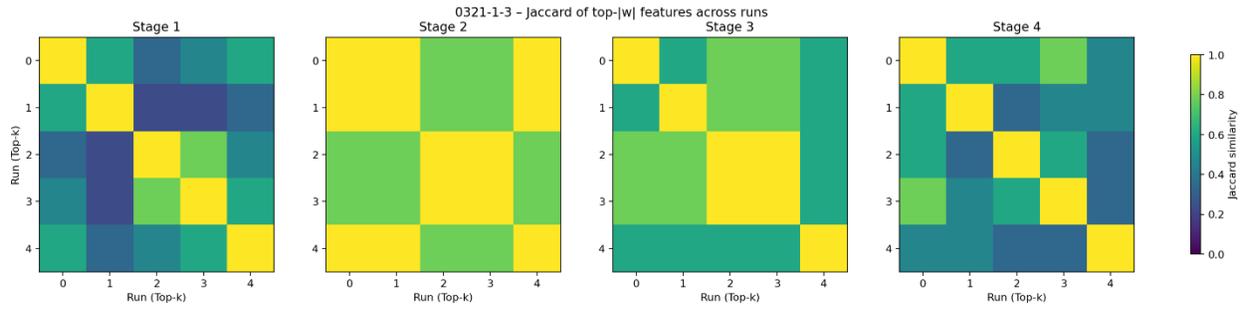

Figure A1 Jaccard Similarity for Subject 0321-1

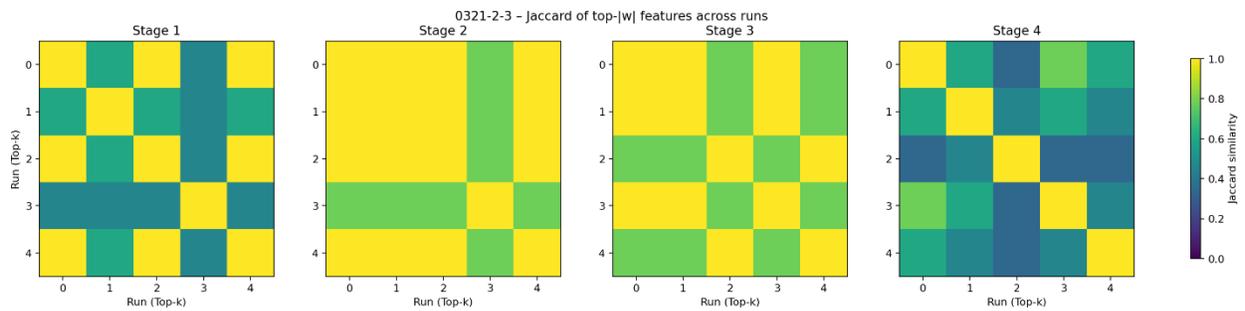

Figure A2 Jaccard Similarity for Subject 0321-2

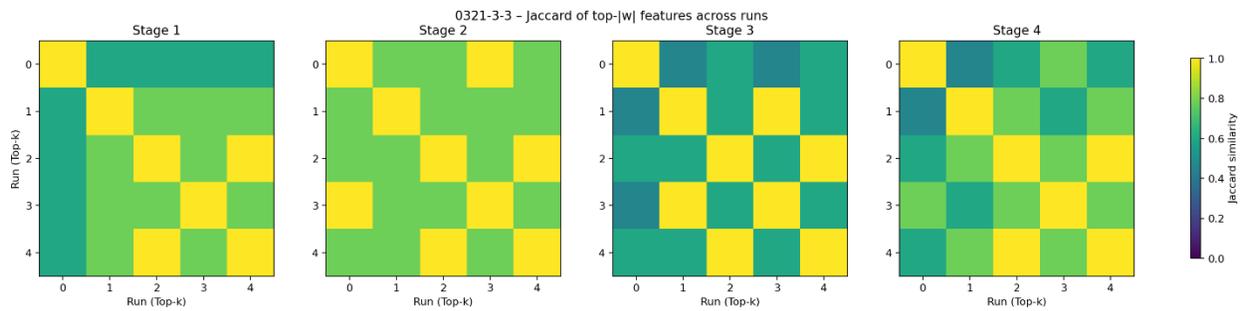

Figure A3 Jaccard Similarity for Subject 0321-3

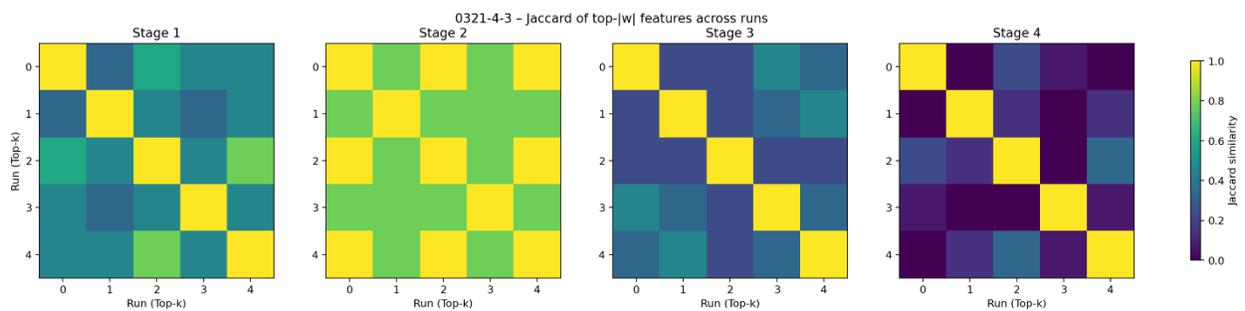

Figure A4 Jaccard Similarity for Subject 0321-4

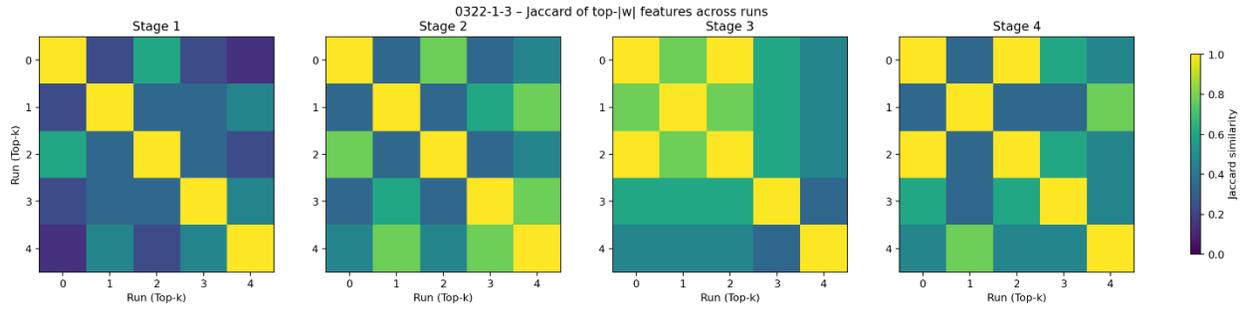

Figure A5 Jaccard Similarity for Subject 0322-1

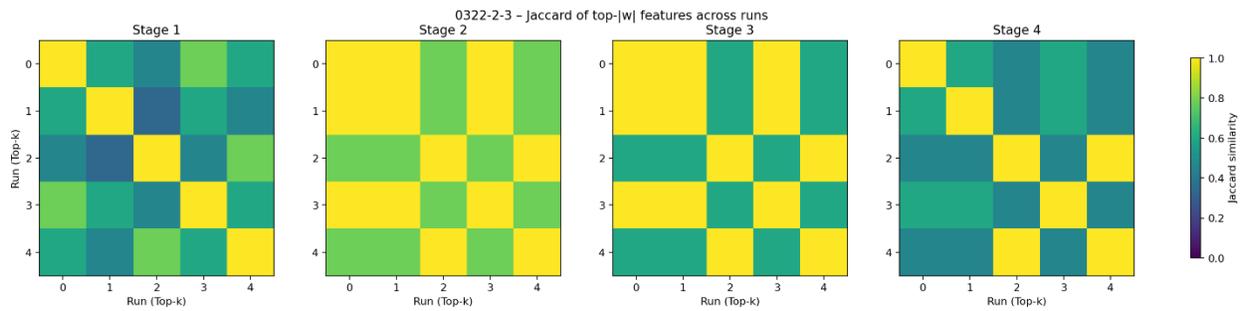

Figure A6 Jaccard Similarity for Subject 0322-2

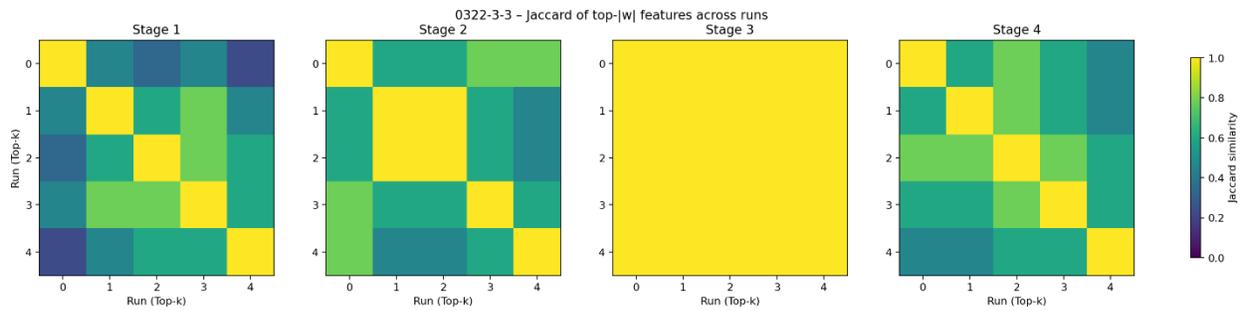

Figure A7 Jaccard Similarity for Subject 0322-3

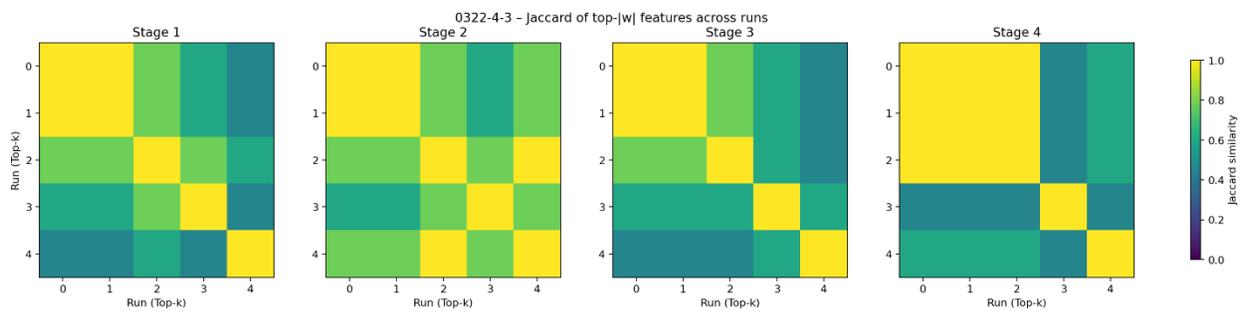

Figure A8 Jaccard Similarity for Subject 0322-4

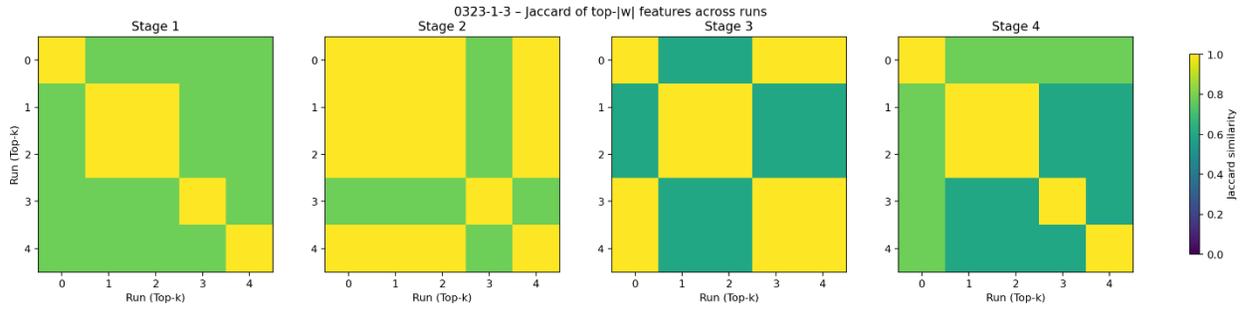

Figure A9 Jaccard Similarity for Subject 0323-1

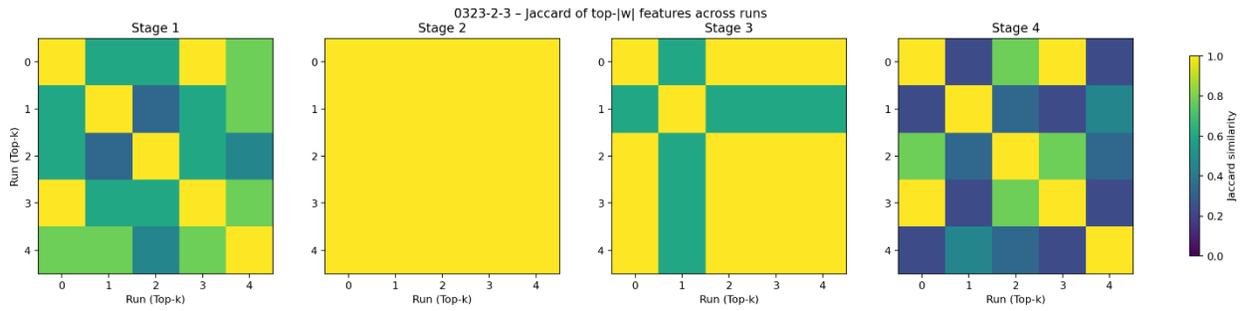

Figure A10 Jaccard Similarity for Subject 0323-2

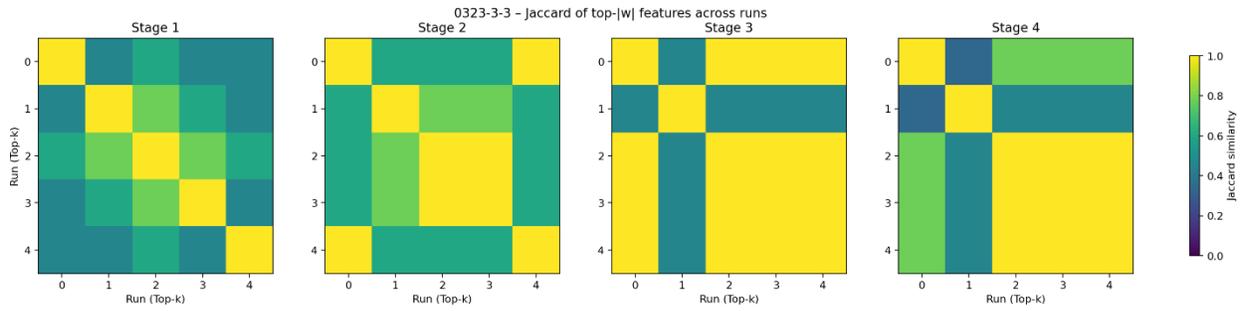

Figure A11 Jaccard Similarity for Subject 0323-3

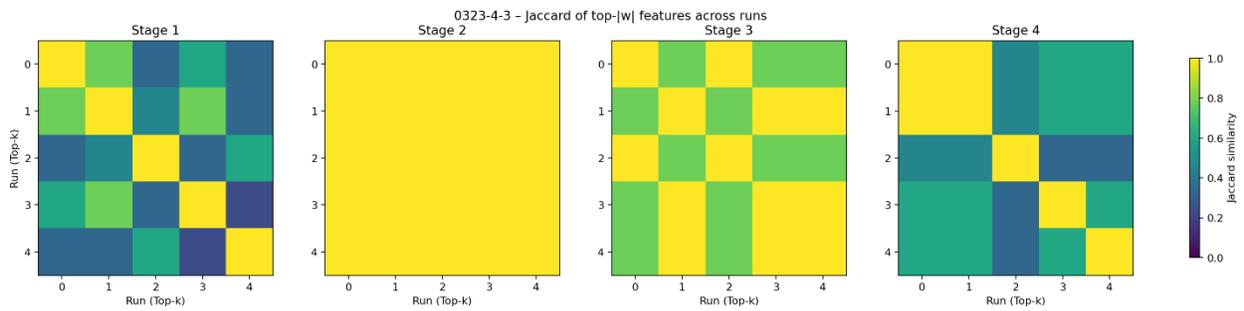

Figure A12 Jaccard Similarity for Subject 0323-4